\documentclass[preprint,12pt]{elsarticle}

\usepackage{amssymb}
\usepackage{amsmath}
\usepackage{algorithm}%
\usepackage{algorithmicx}%
\usepackage{algpseudocode}%
\usepackage{listings}%
\usepackage{multirow}
\usepackage{makecell}
\usepackage{dcolumn}
\usepackage{subfigure}
\usepackage{caption}
\usepackage{graphicx}
\usepackage{multirow}
\usepackage{booktabs}

\journal{Neural Networks}

\begin{document}

\begin{frontmatter}



\title{Curriculum Reinforcement Learning with Measurable Task Representation Learning}


\author{Yongyan Wen $^{1}$, Siyuan Li $^{1,*}$, Mingjian Fu $^{2}$, Yiqin Yang $^{3}$, Xun Wang $^{4}$ and Peng Liu $^{1}$}

\affiliation{organization={Harbin Institute of Technology},
            city={Harbin},
            postcode={150001}, 
            state={Heilongjiang},
            country={China}}

\affiliation{organization={Fuzhou University},
            city={Fuzhou},
            postcode={350108}, 
            state={Fujian},
            country={China}}
            
\affiliation{organization={China Academy of Sciences},
            city={Beijing},
            postcode={100190}, 
            country={China}}

\affiliation{organization={CASIC},
            city={Beijing},
            postcode={100043}, 
            country={China}}

\begin{abstract}
In curriculum reinforcement learning (CRL), an agent incrementally accumulates knowledge over a sequence of tasks (i.e., a curriculum), and the learning process is aimed at using the accumulated knowledge to finally solve a challenging target task. While early CRL works focus on sequencing candidate tasks, recent research explores automatic curriculum generation. Among the rich CRL literature, the interpolation-based CRL paradigm is a main body, which automatically generates intermediate tasks by interpolating between the initial task distribution and the target task distribution in task space with meaningful distance metrics (i.e., can measure the task similarity). However, in challenging navigation tasks, the non-Euclidean context (task) space invalidates this assumption. To achieve automatic curriculum generation in complex task, we propose a novel automatic curriculum generation approach based on measurable task representation learning. To better measure the similarity, we propose to transform the task space to a latent space. Through a variational autoencoder structure that encodes the reward and the state transitions, we achieve a latent task representation with a task similarity measurement property, and two close task embeddings correspond to two similar tasks in terms of rewards and state transitions. Based on the learned task representation, we further develop an automatic curriculum generation scheme, which can effectively generate new tasks more and more similar to the target task. We evaluate our method in a variety of challenging navigation tasks, and the experiment results indicate that the proposed approach surpasses state-of-the-art CRL approaches based on interpolation and generative adversarial networks.
\end{abstract}


\begin{highlights}
\item An automated curriculum reinforcement learning method without external rewards
\item Learning task representations for effective similarity measurement
\item  Mapping tasks to latent space, overcoming Euclidean limitations
\item Generating adaptive curricula for better learning efficiency
\end{highlights}


\end{frontmatter}



\section{Introduction}\label{sec.introduction}
Reinforcement learning (RL) \cite{sutton_reinforcement_2018} has emerged as a promising learning paradigm in challenging sequential decision-making tasks, including applications such as playing video games \cite{mnih_human-level_2015}, chess \cite{silver_general_2018}, and robotic hand control \cite{zhu_dexterous_2019, akkaya_solving_2019, zhuang2023robot, pmlr-v229-defazio23a}. Nevertheless, addressing long-horizon tasks with sparse rewards poses a challenge, as an agent must execute a sequence of correct actions to receive a positive reward, making exploration challenging in practice. For example, in a maze setting, the agent receives a positive reward only upon reaching the target goal. To address this challenge, curriculum reinforcement learning (CRL) \cite{narvekar_curriculum_2020} enables the agent to learn in a sequence of tasks (curriculum) with increasing difficulty, accumulate knowledge in these tasks, and finally solve the challenging target task.

Curriculum learning \cite{bengio_curriculum_2009} accelerates the learning process of difficult tasks by arranging learning samples into a meaningful sequence. For example, in a video game, the player often begins with easy beginner levels. As the game progresses, the levels become progressively more challenging. In contrast to the level sequence with increasing difficulty, attempting to directly confront and overcome the most formidable level can be exceedingly hard. Analogously, in CRL, a policy is initially trained on easy tasks and gradually transfers to more complex ones, incrementally increasing the task difficulty or environmental complexity. As CRL approaches are able to accumulate knowledge in the task sequence and transfer the knowledge to new tasks, the CRL scheme can gradually learn the policy to solve the challenging target task, which is infeasible to solve for the approaches learning from scratch.

\begin{figure}[b]
    \centering
    \subfigure[]{
        \begin{minipage}{0.18\linewidth}
            \centering\includegraphics[width=\linewidth]{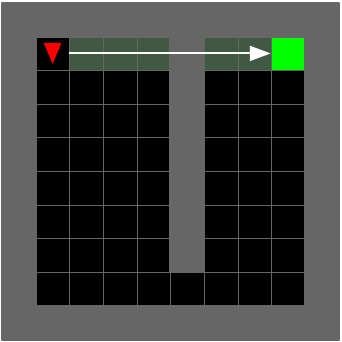}
        \end{minipage}
        \label{fig.motivation_a}
    }
    \subfigure[]{
        \begin{minipage}{0.38\linewidth}
            \centering\includegraphics[width=\linewidth]{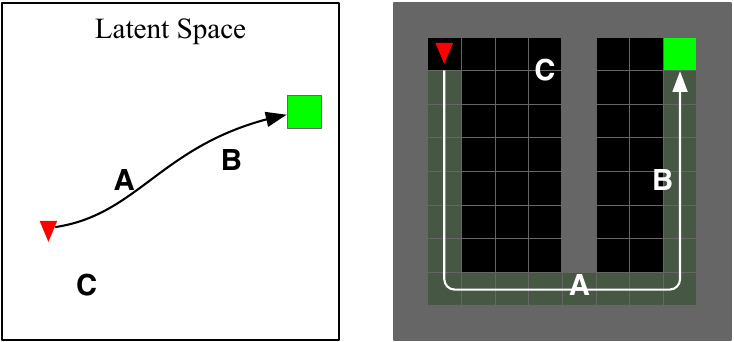}
        \end{minipage}
        \label{fig.motivation_b}
    }
    \caption{Compared to direct interpolation in context space \ref{fig.motivation_a} without considering task similarity metrics. Our desired approach, which involves transforming into latent space \ref{fig.motivation_b} for a more refined task similarity measurement, is capable of generating more efficient intermediate tasks.}
    \label{fig.motivation}
\end{figure}

A crucial problem in CRL is automatic curriculum generation. A desired curriculum can guide the agent to effectively autonomously practice the intermediate tasks and solve the target task, while a poor curriculum may be too easy, too difficult, or unrelated to the target task. However, it is uncertain how to generate the curriculum for a given environment, which has led to various methods aimed at automating this process. A popular automatic curriculum generation paradigm is task distribution interpolation \cite{klink_self_2020a, klink_self_2020b, klink_a_2021, klink_curriculum_2022, huang_curriculum_2022}. The task interpolation-based methods generate new tasks by interpolating between the current task distribution and the target task distribution, and these methods expect that the generated new tasks are more similar to the target task. Specifically, the distance metrics for interpolation are KL divergence \cite{klink_self_2020a, klink_self_2020b} or Wasserstein distance \cite{klink_curriculum_2022, huang_curriculum_2022}. These methods have a great experiment performance when the distance in the context (task) space can measure the task difficulty. However, they assume that the distance between the tasks can be measured by the Euclidean distance metric, which may encounter limitations in the environment with complex geometric structure. For example, as shown in Figure \ref{fig.motivation}, in a simple maze, the straight path from the initial position to the target may be blocked by the wall, thus making intermediate tasks infeasible.

To overcome the above challenge in CRL, we estimate task similarity by transforming the context space into a measurable latent space. We introduce \textbf{A}utomatic \textbf{C}urriculum with \textbf{R}epresentation \textbf{L}earning (ACRL), a novel automatic curriculum generation approach with task representation learning. As a task is formulated with a Markov decision process, we propose to measure the similarity between tasks by the state transition and the reward. Therefore, we propose a variational autoencoder (VAE) \cite{kingma_auto_2013} task representation learning method, which encodes the trajectories collected in different tasks to the latent space as embeddings and the decoder restores the embeddings to the next state and the reward. Regulated by the decoder, the latent space representation (embedding) in the proposed approach is able to measure task similarity. Furthermore, an additional task decoder maps the embeddings back to the context space so that we can generate new tasks and handle both parametric and empirical distributions. By learning with the policy simultaneously, a curriculum is formed by generating intermediate task distributions from the source (initial) distribution to the target. Finally, since our method is used to generate the curriculum for RL agent training, it is compatible with any existing RL algorithms, such as PPO\cite{schulman_proximal_2017} and SAC\cite{soft_haarnoja_2018}. Finally, we note that ACRL assumes a parametric context space, where each context is represented by numerical parameters of the environment. This assumption ensures that contexts can be embedded into the latent space and reconstructed by the task decoder. Our primary contributions are outlined below:

\begin{enumerate}
    \item We propose ACRL, an automated curriculum RL framework that generates intermediate task distributions between the source task distribution and the target. To facilitate the automatic generation of a curriculum and achieve smoother intermediate distributions, we employ task representation to map contexts to the latent space, ensuring a more seamless metric. Our proposed method is able to deal with both discrete and continuous context spaces and does not require external environment reward.
    \item We present a task representation learning method for measuring the task similarity. This method is capable of mapping the context space to the latent space, making it suitable for a wide range of metric spaces, including non-Euclidean metrics.
    \item Through empirical demonstrations, we show that our method has superior learning efficiency and asymptotic performance on various challenging navigation tasks with complex geometry structure when compared to state-of-the-art CRL baselines.
\end{enumerate}

In Section \ref{sec.background}, we provide a concise introduction to the background and notations relevant to our work. Then, in Section \ref{sec.related_work}, we discuss the related works. Subsequently, Section \ref{sec.method} offers a detailed exposition of our methodology and the implementation details. In Section \ref{sec.experiments}, we present the experiment results of our method, substantiating the validity of our method through comparisons with baselines. Finally, in Section \ref{sec.conclusion}, we encapsulate our work, offer conclusions, and address certain limitations and potential avenues for future research.

\begin{figure}[t]
    \centering
    \includegraphics[width=\linewidth]{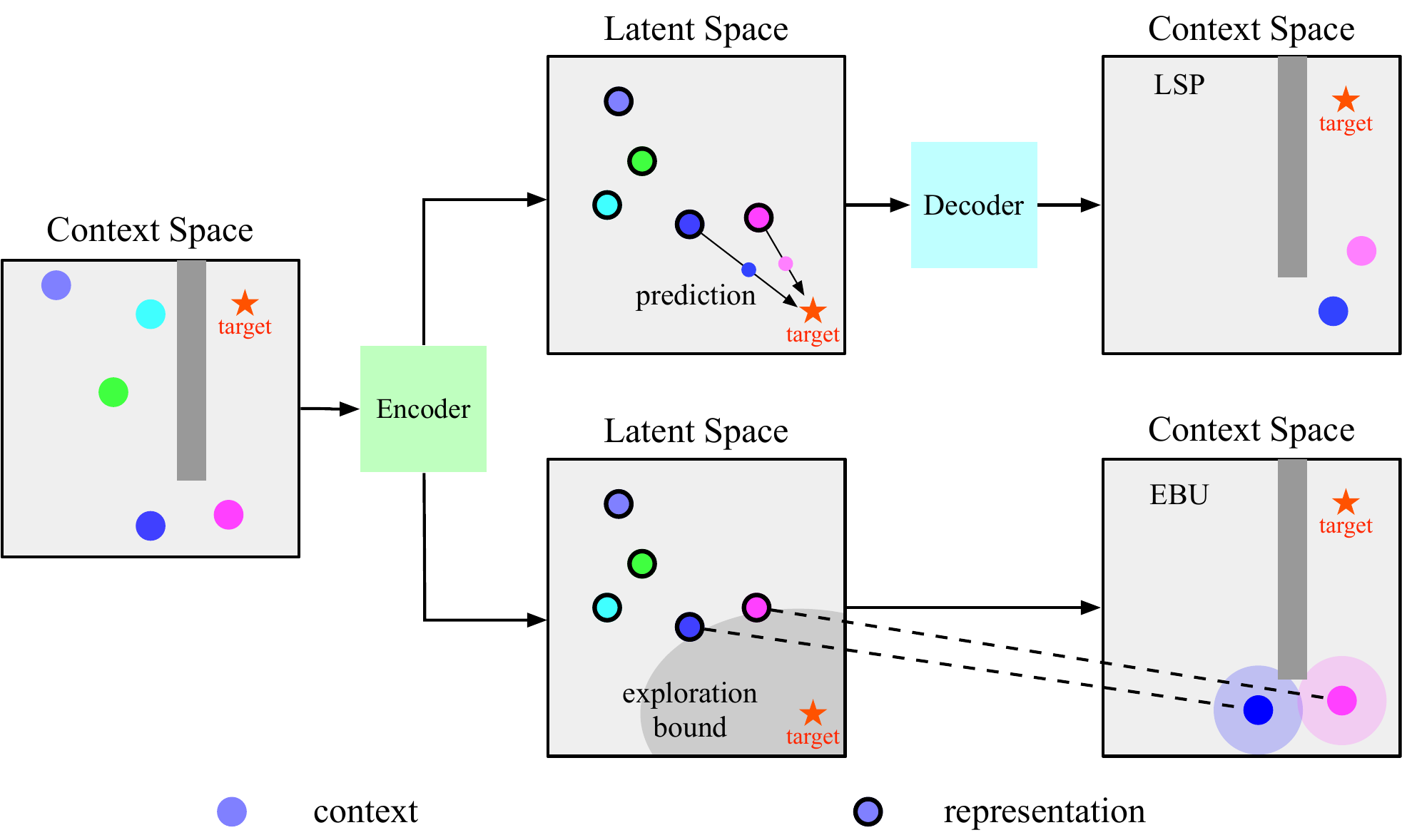}
    \caption{The schema of Latent Space Prediction (LSP) and Exploration Bound Update (EBU). Circles of distinct colors symbolize different contexts, and circles of identical color in the latent space denote corresponding task embeddings. Through task representation learning with the encoder, the context variables are transformed into the latent embeddings in the middle column, yielding a smoother task representation. Leveraging the outcomes of representation learning, LSP performs interpolation in the latent space and subsequently decodes back to the context space. EBU selects updated contexts in the latent space and adds Gaussian noise to generate new tasks.}
    \label{fig.schema}
\end{figure}

\section{Background}\label{sec.background}
Contextual reinforcement learning \cite{hallak_contextual_2015} is an extension of the fundamental RL problem. In comparison to single-task RL, the Contextual Markov Decision Process (CMDP) broadens the scope of the Markov Decision Process (MDP) to encompass multi-task settings. In this study, we focus on a discounted finite-horizon CMDP, which can be represented as a tuple $M=(\mathcal{S}, \mathcal{C}, \mathcal{A}, \mathcal{R}, \mathcal{P}, p_0, \rho, \gamma, H)$. Here, $\mathcal{S}$ is the state space, $\mathcal{A}$ is the action space, $\mathcal{C}$ is the context space, $\mathcal{R}: \mathcal{S}\times \mathcal{A}\times \mathcal{C}\mapsto \mathbb{R}$ is the context-dependent reward space, $\mathcal{P}: \mathcal{S}\times \mathcal{A}\times \mathcal{C}$ is the context-dependent transition function, $p_0:\mathcal{C}\mapsto \Delta(S)$ is the context-dependent initial state distribution, $\rho\in \Delta(\mathcal{C})$ is the context distribution, $\gamma\in (0, 1)$ is the discount factor and $H$ is the horizon. To distinguish between the two important concepts of context and task, we define as follows.

\textit{\textbf{Definition 1 (Context):} A context $c\in \mathcal{C}$ specifies the parameters of a contextual Markov decision process (CMDP), which determine the environment dynamics and reward function. Contexts are sampled from a context space $\mathcal{C}$ and serve as the underlying factors controlling task variation.}

\textit{\textbf{Definition 2 (Task):} A task is an environment instance determined by a specific context $c$. In other words, each task corresponds to a CMDP configured by a particular context.}

Initially, the context $\mathbf{c}$ is sampled from the distribution $\mathbf{c}\sim \rho$ and start from initial state $s_0\sim p_0(\cdot| \mathbf{c})$. At each time step $t$, the agent selects an action by the policy $a_t\sim \pi(s_t, \mathbf{c})$ and receives the reward $\mathcal{R}(s_t, a_t, c)$ from the environment. Subsequently, the environment transits to the next state $s_t'=s_{t+1}\sim \mathcal{P}(\cdot|s_t, a_t, c)$. Thus, a transition can be represented as $\tau_{\mathbf{c},t}=(s_t,a_t,r_t,s_t')$. The process continues until the target is reached or the time step reaches the maximum episode length $H$ and the sampled trajectory is denoted as $\tau_\mathbf{c}=\{(s_t,a_t,r_t,s_t')\}_{t=0}^{H}$. The learning objective is to determine the optimal policy $\pi$ that maximize the expected discounted reward:
\begin{equation}
    \max \mathcal{J}(\pi)_{\mathbf{c}\sim \rho(\mathbf{c})} = \max_\pi \mathbb{E}_{\tau_{\mathbf{c}}} \left[\sum_{t=0}^{H-1} \gamma^t \mathcal{R}(s_t, a_t, \mathbf{c})\right].
\end{equation}

\section{Related Work}\label{sec.related_work}

\subsection{Curriculum Reinforcement Learning}
Curriculum learning aims to enhance the learning efficiency on difficult tasks, and has demonstrated success in various applications \cite{bengio_curriculum_2009, sutskever_on_2013}. In the field of RL, the curriculum improves learning efficiency and helps to tackle more challenging tasks \cite{narvekar_curriculum_2020} and transferring policies to address previously unseen tasks. Some studies focus on sequencing specific pre-designed tasks to construct a curriculum \cite{wu_robust_2022, li_understanding_2022, huang_curriculum_2022}. However, in scenarios where tasks are not predefined, automatic curriculum generation is necessary. Most curriculum RL methods lies in estimating tasks that align with the current policy's capacity boundaries. The estimation approaches can involve temporal steps to reach the goal \cite{campero_learning_2021}, value function \cite{zhang_automatic_2020, chen_variational_2021, kim2023variational} and episodic reward \cite{florensa_reverse_2017, florensa_automatic_2018, racaniere_automated_2019, portelas_teacher_2020}. Some studies have further explored unsupervised environment design with adversarial training \cite{dennis_emergent_2020}, environment design via evolution based on reward \cite{wang_paired_2019} or regret \cite{parker-holder_evolving_2022}, planning reachable goals \cite{zhang_c-planning_2021}, factorizations \cite{mirsky_task_2022}, and employed path planning to generate environments \cite{ao_eat-c_2022}.

A prevalent line in CRL research has framed the problem of automatic curriculum generation as an interpolation of distributions, utilizing Kullback-Leibler (KL) divergence \cite{klink_self_2020a, klink_self_2020b, klink_a_2021} or Wasserstein distance \cite{klink_curriculum_2022, huang_curriculum_2022, cho_outcome-directed_2023} between distributions to generate intermediate task distributions. However, these methods are often limited in capturing task similarity, especially when task distributions are complex or the context space lacks a well-defined metric structure. To address this limitation, we propose a novel CRL approach with task representation learning, where latent embeddings are learned to measure the similarity between tasks.

\subsection{Task Representation Learning}
Utilizing task representations to automate task generation stands out as a viable approach for achieving automated curriculum generation effectively. Some studies have explored the similarities between tasks (MDPs) from the perspective of bisimulation \cite{castro_scalable_2020, zhang_learning_2021}. Employing trajectory embeddings sampled from tasks as representations is a common practice \cite{wang_robust_2017, zintgraf_varibad_2020, rakelly_efficient_2019, jabri_unsupervised_2019}. Wang et al. \cite{wang_robust_2017} leverage the VAE \cite{kingma_auto_2013} with a bidirectional Long Short-Term Memory (LSTM) \cite{graves_bidirectional_2005} to encode trajectories, generating an embedding vector for the policy. Zintgraf et al. \cite{zintgraf_varibad_2020} learn a low-dimensional stochastic latent variable as representation of each MDP in meta RL. Rakelly et al. \cite{rakelly_efficient_2019} encode a sampled trajectory in each task to a Gaussian distribution, creating a permutation-invariant representation. Jabri et al. \cite{jabri_unsupervised_2019} adopt the optimization scheme of DeepCluster \cite{caron_deep_2018} to obtain a trajectory-level task representation. Another study encodes state-action pairs sampled in a set of diverse environments by the policy to describe the dynamics \cite{raileanu_fast_2020}.

Previous works primarily focus on rapid adaptation to new environments with learned representations. In CRL, \cite{fang_adaptive_2020} utilizes GAN to learn task parameters as representations to update the curriculum. \cite{pmlr-v202-azad23a} encodes tasks as sequences of integers and uses LSTM-based Recurrent VAE \cite{bowman2016generating} to learn a latent task manifold. In this paper, we obtain task representations from the perspective of trajectories, but we employ them to describe task similarity and facilitate the prediction of new tasks. To enhance the smoothness of intermediate task distributions, we propose incorporating task representation into curriculum generation. Experiment results demonstrate that the representation-based approach yields desired intermediate task distribution outcomes.

\section{Curriculum Generation with Task Representation Learning}\label{sec.method}
In this paper, we aim to construct intermediate task distributions from the initial task distribution and the target. Section \ref{ssec.task_representaion_learning} describes the techniques used for task representation learning. Section \ref{ssec.automatic_curriculum_generation} then presents our automatic curriculum generation method, which is based on this representation learning. Finally, Section \ref{ssec.implementation_details} provides implementation details of the method.

\subsection{Task Representation Learning}\label{ssec.task_representaion_learning}
For two tasks, denoted as context $\mathbf{c}_1$ and context $\mathbf{c}_2$, the Euclidean distance between them ($\Vert \mathbf{c}_1 - \mathbf{c}_2 \Vert_2$) may fail to capture their true similarity if their trajectories differ significantly under a given policy. Intuitively, task similarity depends on the policy’s behavior, which evolves during learning. To capture this dynamic, we must consider trajectory information—specifically, sequences of states and rewards—which are key characteristics of a Markov Decision Process (MDP) under context $\mathbf{c}$.  

To address this, we propose a VAE-based task representation learning approach that constructs a latent space reflecting task similarity through sampled trajectories. The encoder maps contexts into this latent space, enabling more meaningful similarity metrics than those computed directly in the context space. By generating latent representations for intermediate tasks and reconstructing them into the context space via the decoder, we obtain smoother task distributions, facilitating the construction of a curriculum that enhances policy transfer.

\begin{figure}[t]
    \centering
    \includegraphics[width=0.8\columnwidth]{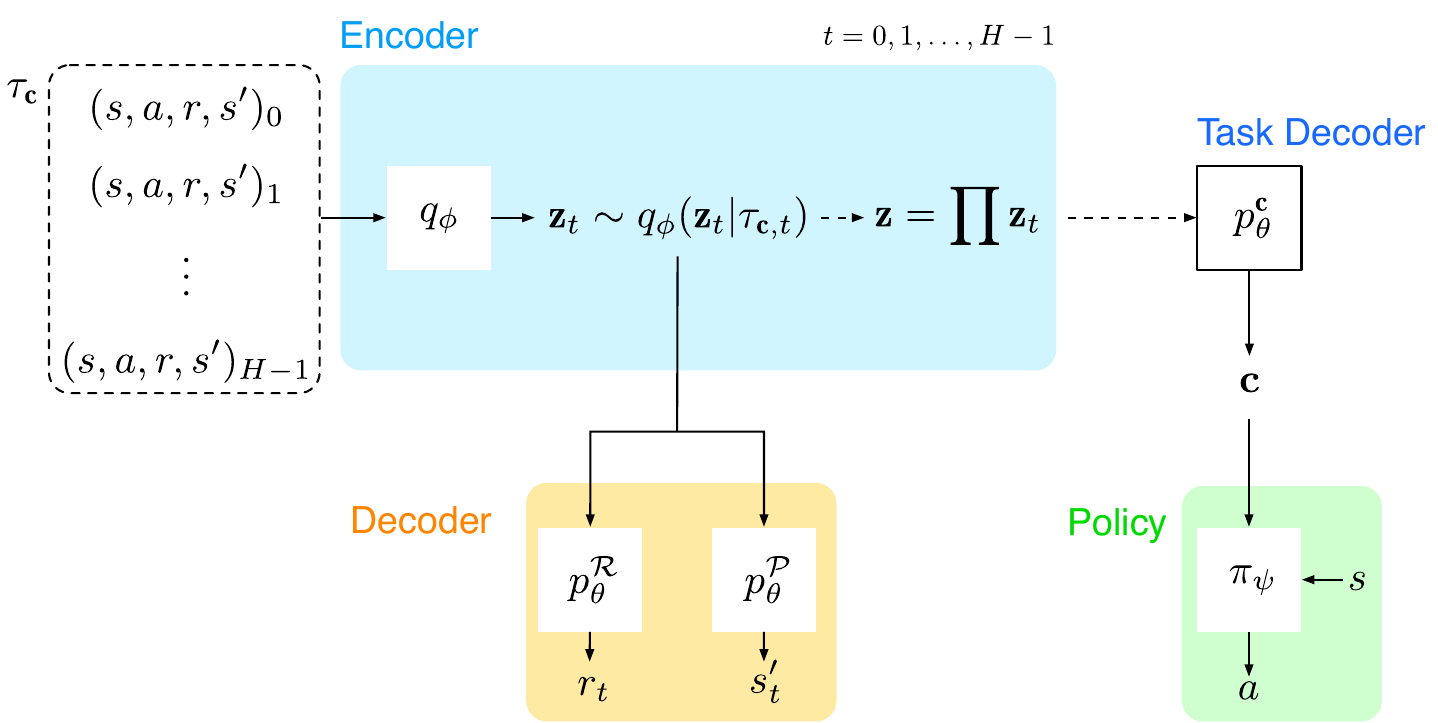}
    \caption{Task representation learning architecture. Representation learning and policy learning are coupled. A trajectory of states and rewards is encoded to produce the posterior as transition embeddings $q_\phi(\mathbf{z}_t|\tau_{\mathbf{c},t})_{t=0,\dots,H-1}$. The representation of the task is achieved by multiplying all transition embeddings within a trajectory. Policy learning is co-trained with representation learning using RL. The dotted line indicates that the part is not backpropagated when updating the parameters.}
    \label{fig.architecture}
\end{figure}

To obtain the latent representation of a task under context $\mathbf{c}$—sampled from a task distribution—we execute the policy and collect a trajectory $\tau_{\mathbf{c}} = \{(s_t, a_t, r_t, s_t')\}_{t=0}^{H-1}$. Here, $H$ denotes the (variable) length of the trajectory. As shown in Figure \ref{fig.architecture}, the input to the VAE’s transition encoder is a single transition $(s_t, a_t, r_t, s_t')$, and its output is a Gaussian distribution $q_\phi(\mathbf{z} \mid \tau_{\mathbf{c},t}) = \mathcal{N}(\mu, \sigma^2 \mathbf{I})$. Latent variables are then sampled as $\mathbf{z}_t \sim q_\phi(\mathbf{z} \mid \tau_{\mathbf{c},t})$.

Rewards and dynamics differ across tasks, but their underlying structure is shared and represented by a context vector $\mathbf{c}$. Thus, sampling from the context distribution $\rho(\mathbf{c})$ is equivalent to sampling from the joint distribution over reward and transition functions, denoted $\rho(\mathcal{R}_\mathbf{c}, \mathcal{P}_\mathbf{c})$, where $\mathcal{R}_\mathbf{c}$ is the reward function and $\mathcal{P}_\mathbf{c}$ is the state transition dynamics of the MDP under context $\mathbf{c}$. Since the context parameters determine both the reward function and the dynamics, the decoder should be able to reconstruct them from the latent representation. The decoder consists of two components: a transition decoder $p^\mathcal{T}_\theta (s'|\mathbf{z})$ and a reward decoder $p^\mathcal{R}_\theta (r|\mathbf{z})$. The reward decoder maps the latent variable $\mathbf{z}$ to the reward $r$, while the transition decoder predicts the next state $s'$. Our goal is to optimize the evidence lower bound:
\begin{equation}\label{eq.vae_loss}
    \begin{aligned}
        \mathbb{E}_{\mathbf{c}\sim \rho(\mathbf{c}), \tau_{\mathbf{c},t}\sim \tau_{\mathbf{c}}}[\log p_\theta(t)]&\geq \text{ELBO}\\
        &=\mathbb{E}_\rho[\mathbb{E}_{q_\phi(\mathbf{z}|\tau_{\mathbf{c},t}))}[\log p_\theta(\tau_{\mathbf{c},t}|\mathbf{z})] - D_{\text{KL}}(q_\phi(\mathbf{z}|\tau_{\mathbf{c},t})\Vert p(\mathbf{z}))]
    \end{aligned}
\end{equation}
where $p_\theta(\tau_{\mathbf{c},t} \mid \mathbf{z}) = p^\mathcal{R}_\theta (r \mid \mathbf{z}) + p^\mathcal{T}_\theta (s' \mid \mathbf{z})$ is the reconstruction loss and $p(\mathbf{z})$ is the prior distribution over the latent variable $\mathbf{z}$. The trajectory representation is computed by the trajectory encoder $q_\phi(\mathbf{z}|\mathbf{c})$.  

The transition and reward functions can be reconstructed from an unordered set of transitions, because such a set is sufficient to train a value function or infer the task \cite{rakelly_efficient_2019}. Therefore, we do not need to model the temporal order of transitions and can achieve a permutation-invariant task representation. To accomplish this, the trajectory embedding is computed as the product of all transition embeddings within the trajectory. Specifically, the task representation is obtained by multiplying the Gaussian distributions of the latent variables corresponding to all transitions in the trajectory:
\begin{equation}\label{eq.task_encoder}
    q_\phi(\mathbf{z}|\tau_\mathbf{c}) = \prod_{t=0}^{H-1} q_\phi (\mathbf{z}_t|\tau_{\mathbf{c},t})
\end{equation}
where $q_\phi(\mathbf{z}_t|\tau_{\mathbf{c},t})_{t=0, \dots, H-1}$ is the distribution output of encoder.

Finally, we introduce an auxiliary task decoder $p_\theta^{\mathbf{c}}(\mathbf{z})$, which differs from the reward and transition decoders in the VAE. While the VAE ensures that latent variables capture information about the reward and dynamics, this auxiliary decoder explicitly maps the latent representation back to the context $ \mathbf{c}$. This mapping is essential for generating new tasks by interpolating or extrapolating in the latent space.

Formally, given a latent variable $\mathbf{z}$, the task decoder predicts the corresponding context $\hat{\mathbf{c}}=p_\theta^\mathbf{c}(\mathbf{z})$, and its objective is to minimize the context reconstruction error:
\begin{equation}\label{eq.task_decoder}
    \mathcal{L}(\theta)=\Vert \mathbf{c}-p_\theta^\mathbf{c}(\mathbf{z})\Vert_2.
\end{equation}

Note that not all trajectories can be represented meaningfully. Since the under-trained policy may perform random actions and fail tasks, we only select trajectories with episodic return exceeding the return threshold $\delta$ for training to ensure accurate evaluation. The process is represented as shown in Algorithm \ref{algo.representation}.

\begin{algorithm}[t]
    \caption{Task Representation Learning}
    \label{algo.representation}
    \begin{algorithmic}[1]
        \State \textbf{Input:} return threshold $\delta$, VAE buffer $\mathcal{B}$ collected on $p(\textbf{c})$

        \State Sample episodes $\{\tau_i|G^\pi (\tau_i)\geq \delta\}\sim \mathcal{B}$

        \If {$\{\tau_i\}$ is not empty}
            \State Update encoder $q_\phi$ and decoder $p_\theta^\mathcal{R},p_\theta^\mathcal{P}$ (Eq. \ref{eq.vae_loss})
            \State Update task decoder $p_\theta^\mathbf{c}$ separately (Eq. \ref{eq.task_decoder})
        \EndIf
        \State \textbf{Output:} $q_\phi, p_\theta^\textbf{c}$
    \end{algorithmic}
\end{algorithm}

\subsection{Automatic Curriculum Generation}\label{ssec.automatic_curriculum_generation}
Using task representations, intermediate task distributions are generated by predicting latent variables that are closer to the target task in the latent space and mapping them back to the context space. As the task distribution is continuously updated, these distributions form a curriculum. Initially, the policy is trained on tasks sampled from the initial distribution, and the resulting experiences are used to train the VAE. The task update algorithm then refines the task distribution based on the policy’s learning progress and the latent representations of tasks. Subsequent training uses tasks sampled from the updated distribution.

During the initial phase, the agent lacks an effective policy for the given tasks, so the task representations are not yet meaningful enough for immediate use. To regulate learning progress, a threshold—such as the average return during evaluation—can be set to determine whether the policy is sufficiently trained on the current task distribution. Once the policy reaches this threshold, the VAE synchronizes task representation updates and refines the task distribution, leveraging its representational and predictive capabilities to guide learning effectively.

We propose two methods for updating the task distribution based on learning progress. These two methods are not used simultaneously; instead, the choice depends on the training phase and the accuracy of the task representation (Section \ref{ssec.implementation_details}). The first method is Latent Space Prediction (\textbf{LSP}). Using the output of the task decoder, we predict the distribution of tasks that are close to the target context $\mathbf{c}_{\text{tar}}$ in the latent space and map it back to the context space to update the task distribution. The second method is Exploration Bound Update (\textbf{EBU}). It predicts tasks in the latent space near the current exploration boundary, treating contexts in this region as those closest to the target among the tasks learned so far. Contexts from this region are sampled and perturbed with noise in the context space to form the new task distribution.

Additionally, we introduce an sampling ratio $\lambda \in [0,1]$ to control the sampling process: with probability $\lambda$, a context is sampled from the task buffer $\mathcal{T}$; otherwise, it is sampled from the uniform distribution $\mathcal{U}(\mathcal{C})$ (see details in Section \ref{ssec.implementation_details}). The overall process is summarized in Algorithm \ref{algo.curriculum_generation}.

\begin{algorithm}[]
    \caption{\textbf{A}utomatic \textbf{C}urriculum with \textbf{R}epresentation \textbf{L}earning (\textbf{ACRL})}
    \label{algo.curriculum_generation}
    \begin{algorithmic}[1]
        \State \textbf{Input:} source task distribution $p(\mathbf{c})$, uniform distribution $\mathcal{U}(\mathbf{c})$, target task $\mathbf{c}_{\text{tar}}$, policy $\pi_\psi$, train steps $N$, update threshold $G$, VAE buffer $\mathcal{B}$, sampling ratio $\lambda$, return threshold $\delta$, target thershold $\delta_{\text{tar}}$, update frequency $n_u$
        \State Current train steps $n\leftarrow 0$, episode $u\leftarrow 0$
        \While {$n<N$}
            \State Sample task $\mathbf{c}\sim p(\mathbf{c})$ with probability $\lambda$, otherwise $\mathbf{c}\sim \mathcal{U}(\mathbf{c})$
            \State Current episode $\tau\leftarrow \{\}$
            \For {$t=0, \dots, H-1$}
                \State Roll out policy on current task to get transition $(s_t,a_t,r_t,s_{t}')$
                \State $\tau \leftarrow \tau\cup(s_t,a_t,r_t,s_t')$
            \EndFor
            \State $n\leftarrow n+H$
            \State $\mathcal{B}\leftarrow \mathcal{B}\cup (\mathbf{c},\tau)$
            \State Update policy $\pi_\psi(s,\mathbf{c})$
            \If {$u\mod n_u==0$}
                \State Sample $m$ trajectories from $\mathcal{B}$, evaluate average episodic return $\bar{G}=\frac{1}{m}\sum_{i=1}^m \sum_{j=0}^{H-1} r_{i,j}$
                \If {$\bar{G} > G$}
                    \State Update task representation (Algorithm \ref{algo.representation})
                    \State Sample $k$ trajectories from target task $\mathbf{c}_{\text{tar}}$, evaluate average episodic return $\bar{G}_{\text{tar}}$
                    \If {$\bar{G}_{\text{tar}} < \delta_{\text{tar}}$}
                        \State Update task distribution $p(\mathbf{c})$ with EBU (Algorithm \ref{algo.ebu})
                    \Else
                        \State Update task distribution $p(\mathbf{c})$ with LSP (Algorithm \ref{algo.lsp})
                    \EndIf
                \EndIf
            \EndIf
            \State $u\leftarrow u+1$
        \EndWhile
        \State \textbf{Output:} policy $\pi_\psi$
    \end{algorithmic}
\end{algorithm}

\begin{algorithm}[t]
    \caption{\textbf{L}atent \textbf{S}pace \textbf{P}rediction (\textbf{LSP})}
    \label{algo.lsp}
    \begin{algorithmic}[1]
        \State \textbf{Input:} task-trajectory pairs $(\mathbf{c}_0,\tau_{\mathbf{c}_0}),\dots,(\mathbf{c}_{n-1},\tau_{\mathbf{c}_{n-1}})$ from the VAE buffer, interpolation parameter $\alpha$, return threshold $\delta$
        \State Updated tasks ${\mathbf{c}}\leftarrow \{\}$
        \State Filter out tasks with low returns: $\{\mathbf{c}_i\}\leftarrow \{G^\pi (\tau_{\mathbf{c}_i})\geq \delta\}$
        \State Calculate the latent space embeddings $\mathbf{z}_0,\dots,\mathbf{z}_{n-1}$ for each task (Eq. \ref{eq.task_encoder})
        \State $\mathbf{z}_{\text{tar}}\leftarrow q_\phi (\mathbf{z}|\tau)$
        \State $\{\mathbf{z}_i'\}\leftarrow \{\mathbf{z}_i'|\mathbf{z}_i'=\alpha \mathbf{z}_{\text{tar}}+(1-\alpha)\mathbf{z}_i\}_{i=1,\dots,n-1}$
        \For {$i=0,\dots,n-1$}
            \State $\{\mathbf{c}\}\leftarrow \{\mathbf{c}\}\cup p_\theta^\mathbf{c}(\mathbf{z}_i')$
        \EndFor
        \State \textbf{Output:} updated tasks $\{\mathbf{c}_i\}$
    \end{algorithmic}
\end{algorithm}

\begin{algorithm}[t]
    \caption{\textbf{E}xploration \textbf{B}ound \textbf{U}pdate (\textbf{EBU})}
    \label{algo.ebu}
    \begin{algorithmic}[1]
        \State \textbf{Input:} task-trajectory pairs $(\mathbf{c}_0,\tau_{\mathbf{c}_0}),\dots,(\mathbf{c}_{n-1},\tau_{\mathbf{c}_{n-1}})$ from the VAE buffer, distribution parameter $\beta$, return threshold $\delta$
        \State Updated tasks ${\mathbf{c}}\leftarrow \{\}$
        \State Filter out tasks with low returns: $\{\mathbf{c}_i\}\leftarrow \{G^\pi (\tau_{\mathbf{c}_i})\geq \delta\}$
        \State Sample the trajectory $\tau_{\mathbf{c}_{\text{tar}}}$ from target task $\mathbf{c}_{\text{tar}}$ with the highest episodic return
        \State $\mathbf{z}_{\text{tar}}\leftarrow q_\phi (\mathbf{z}|\tau_{\mathbf{c}_{\text{tar}}})$
        \State Calculate the latent space embeddings $\mathbf{z}_0,\dots,\mathbf{z}_{n-1}$ for each task (Eq. \ref{eq.task_encoder})
        \State Sort $\mathbf{c}_0,\dots,\mathbf{c}_{n-1}$ in decending order according to the distance $d(\mathbf{c},\mathbf{c}_{\text{tar}})$
        \For {$i=0,\dots,n-1$}
            \State $\{\mathbf{c}\}\leftarrow \{\mathbf{c}\}\cup (\mathbf{c}_{\xi_i}+\mathcal{N}(\mu,\sigma^2)), \xi_i\sim P(\xi;\beta)$ and $\mathbf{c}_{\xi_i}$ is the original task corresponding to $\mathbf{z}_{\xi_i}$
        \EndFor
        \State \textbf{Output:} updated tasks $\{\mathbf{c}\}$
        \end{algorithmic}
\end{algorithm}

\subsubsection{Latent Space Prediction}
For latent space prediction, since the latent space metric reflects task similarity, we can interpolate task embeddings in the latent space and map them back to their corresponding tasks in the original context space. As depicted in Figure \ref{fig.schema}, the target task is $\mathbf{c}_{\text{tar}}$, its latent space representation is $\mathbf{z}_{\text{tar}}$, and tasks are sampled from the current distribution of tasks $\mathbf{c}\sim p(\mathbf{c})$. Given that only tasks leading to the target task yield a meaningful representation, we directly retrieve $n$ context–trajectory pairs $\{(\mathbf{c}_i,\tau_{\mathbf{c}_i})\}_{i=0,\dots,n-1}$ from the VAE buffer, which stores past experiences. From these pairs, we select those whose trajectories achieve episodic returns above the threshold: $\{{\mathbf{c}_i|G^\pi (\tau_{\mathbf{c}_i}) \geq \delta}\}$, thereby avoiding additional rollouts. These trajectories are then encoded to derive the latent variable $\mathbf{z}_i=q_\phi(\mathbf{z}_i|\tau_{\mathbf{c}_i})$. To prioritize tasks closer to the target for prediction, the distance $d(\mathbf{c},\mathbf{c}_{\text{tar}})$ between $\mathbf{z}$ and $\mathbf{z}_{\text{tar}}$ is utilized to rank the tasks in the latent space:
\begin{equation}\label{eq.distance}
    d(\mathbf{c}, \mathbf{c}_{\text{tar}})=\Vert \mathbf{z}-\mathbf{z}_{\text{tar}}\Vert_2.
\end{equation}

Sorting results in a sequence of tasks arranged from smallest to largest distances to the latent variables of the target task. Next, the latent space representation of the predicted tasks is obtained through linear interpolation in the direction of the target task's latent variable.
\begin{equation}\label{eq.latent_interpolation}
    \mathbf{z}'=\alpha \mathbf{z}_{\text{tar}}+(1-\alpha)\mathbf{z}.
\end{equation}
where $\mathbf{z}'$ is the latent space representation of the updated task and $\alpha$ is the adjustable interpolation parameter. Finally, the task decoder is utilized to map back to the context space to obtain the updated task: $\mathbf{c}'=p_\theta^\mathbf{c}(\mathbf{z}')$. 

Note that LSP requires at least one successful trajectory on the target task to provide a meaningful latent embedding; in extremely sparse-reward environments where such trajectories are unavailable, curriculum updates fall back to EBU alone. The overall process is summarized in Algorithm \ref{algo.lsp}.

\subsubsection{Exploration Bound Update}
An alternative method for updating the task distribution involves estimating the current exploration boundary. This is achieved by identifying tasks in the proximity of the boundary and introducing suitable noise to the context space to generate the updated tasks. The explored region encompasses tasks reachable by the current policy $\pi_\psi$, with a round return exceeding the threshold $\delta$. The bound of this region is termed the exploration bound, as depicted in Figure \ref{fig.schema}.

Similar to LSP, once the policy and representation learning are sufficiently trained, we retrieve $n$ task–trajectory pairs $\{(\mathbf{c}_i,\tau_{\mathbf{c}_i})\}_{i=1,\dots,n}$ directly from the VAE buffer, which maintains previously collected experiences. Among these, we retain only the tasks whose trajectories achieve episodic returns above the threshold $\{\mathbf{c}_i|G^\pi(\tau_{\mathbf{c}_i})\geq \delta\}$. The corresponding trajectories are then encoded to obtain embeddings $\mathbf{z}_i = q_\phi(\mathbf{z}_i|\tau_{\mathbf{c}_i})$. To facilitate the selection of the nearest embedding, we sort the latent variables based on Euclidean distances (Eq. \ref{eq.distance}) to the target latent variables $\mathbf{z}_{\text{tar}}$. The task sequence $\mathbf{z}_1, \mathbf{z}_2, \dots, \mathbf{z}_n$ is sorted in descending order of their distances, yielding the corresponding sequence of tasks $\mathbf{c}_1, \mathbf{c}_2, \dots, \mathbf{c}_n$. To augment the diversity of generated tasks and enhance the generalization capability of the RL policy, the index $\xi$ is sampled from the sequence using an exponential distribution with parameter $\beta$:
\begin{equation*}
    P(\xi;\beta)= \beta e^{-\beta \xi}, \xi\geq 0,
\end{equation*}
where the index $\xi$ requires a rounding operation $\text{round}(\xi)$. We sample $m$ tasks $\{\mathbf{c}_{\xi_i}|\xi_i\sim P(\xi;\beta)\}_{i=1,\dots,m}$ as the bound tasks. Finally, add a dimensionally independent Gaussian noise $\mathcal{N}(0, \sigma^2 \mathbf{I})$ as a new context of the task instance. In cases where the context space is discrete, discretization of the result is necessary because the output of the decoder is continuous. The overall procedure is shown in Algorithm \ref{algo.ebu}.

\subsection{Implementation Details}\label{ssec.implementation_details}
In the actual implementation, we use a combination of both update methods. Initially, the representation of the target task is inaccurate because the policy struggles to obtain a successful trajectory on the target; at this stage, we use \textbf{EBU} to estimate the boundary task of the current policy. When the episodic return of trajectories sampled on the target reaches the target threshold $\delta_{\text{tar}}$, the method switches to \textbf{LSP} to further approximate the target.

The intermediate task distributions are nonparametric and represented as sets of context samples. To simplify task sampling and the fusion of different distributions with varying weights, we maintain a task buffer $\mathcal{T}$ that stores sampled tasks. Furthermore, to improve exploration and avoid low diversity in the generated tasks, we do not sample exclusively from $\mathcal{T}$, but instead mix it with a uniform distribution over the context space using a sampling ratio $\lambda$. Denoting the context space of the environment as $\mathcal{C}$, the uniform distribution is $\mathcal{U}(\mathcal{C})$. During context sampling, we draw a context uniformly from $\mathcal{T}$ with probability $\lambda$, or from $\mathcal{U}(\mathcal{C})$ with probability $1 - \lambda$.

Trajectories with episodic returns greater than a return threshold $\delta$—along with their corresponding tasks—are selected from the VAE buffer $\mathcal{B}$, which is first-in-first-out and maintains a fixed size. Specifically, we select $\{\tau_{\mathbf{c}_i} | G^\pi(\tau_{\mathbf{c}_i}) \geq \delta\}$. These trajectories are then used to update the latent variables (Eq. \ref{eq.latent_interpolation}), and the resulting updated tasks are added to $\mathcal{T}$ to form the updated task distribution.

\section{Experiments}\label{sec.experiments}
In this section, we aim to address the following questions by evaluating several experiments: 1) Can ACRL improve the learning efficiency of reinforcement learning? 2) How does ACRL compare to the existing baseline algorithms in terms of improving RL performance? 3) What task distribution of curriculum can be generated by ACRL during training?

To address the first and second questions, we conducted evaluations of our algorithm on various widely used RL tasks and compared its performance to several advanced CRL algorithms. The \texttt{MiniGrid} domain, a well-established benchmark in RL research \cite{MinigridMiniworld23}, serves as a demonstration of our algorithm's ability to enhance learning efficiency on the target task through the creation of a guided curriculum. Additionally, for the continuous control task, we introduce the \texttt{U-Maze} environment for further testing. Concerning the third question, we provide evidence that ACRL robustly generates guided task distributions.

\subsection{Experiment Setup}

\begin{figure}[t]
    \centering
    \subfigure[Easy-A]{
        \begin{minipage}{0.22\linewidth}
            \centering\includegraphics[width=\linewidth]{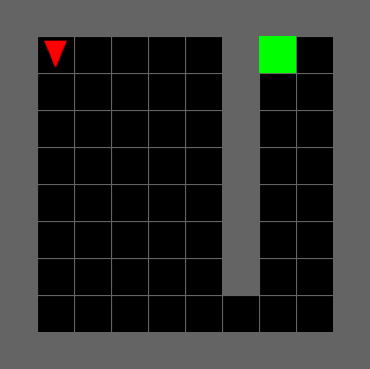}
        \end{minipage}
        \label{fig.minigrid_env_e_a}
    }
    \subfigure[Easy-B]{
        \begin{minipage}{0.22\linewidth}
            \centering\includegraphics[width=\linewidth]{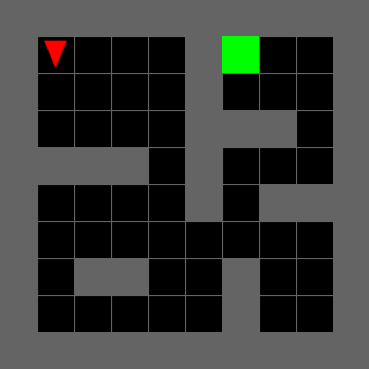}
        \end{minipage}
        \label{fig.minigrid_env_e_b}
    }
    \subfigure[Easy-C]{
        \begin{minipage}{0.22\linewidth}
            \centering\includegraphics[width=\linewidth]{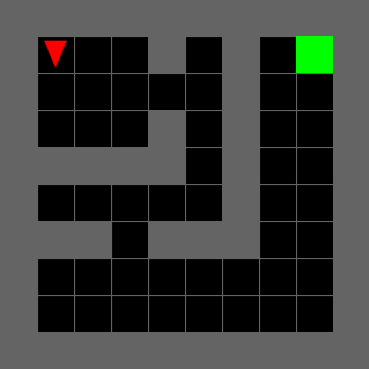}
        \end{minipage}
        \label{fig.minigrid_env_e_c}
    }
    \subfigure[Medium-A]{
        \begin{minipage}{0.22\linewidth}
            \centering\includegraphics[width=\linewidth]{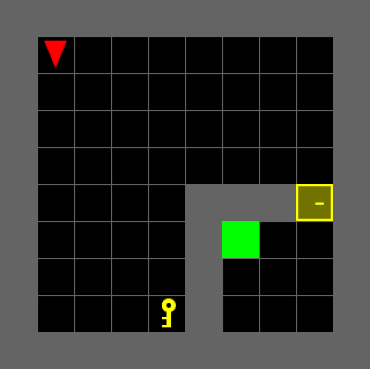}
        \end{minipage}
        \label{fig.minigrid_env_m_a}
    }
    \subfigure[Medium-B]{
        \begin{minipage}{0.22\linewidth}
            \centering\includegraphics[width=\linewidth]{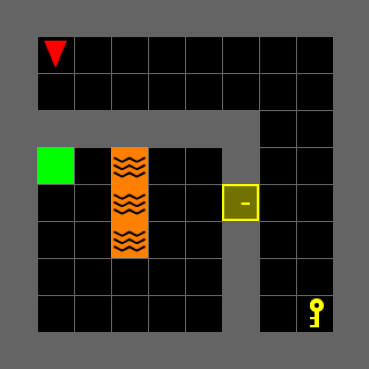}
        \end{minipage}
        \label{fig.minigrid_env_m_b}
    }
    \subfigure[Medium-C]{
        \begin{minipage}{0.22\linewidth}
            \centering\includegraphics[width=\linewidth]{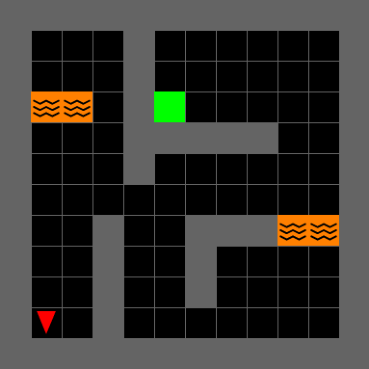}
        \end{minipage}
        \label{fig.minigrid_env_m_c}
    }
    \subfigure[Hard-A]{
        \begin{minipage}{0.22\linewidth}
            \centering\includegraphics[width=\linewidth]{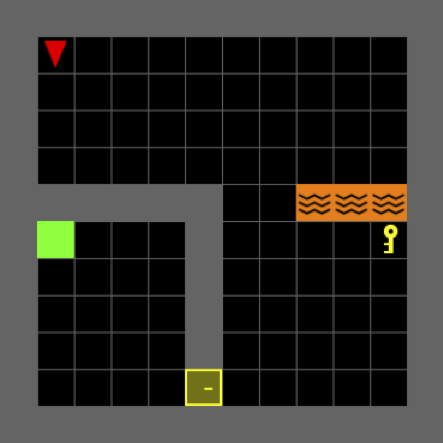}
        \end{minipage}
        \label{fig.minigrid_env_h_a}
    }
    \subfigure[Hard-B]{
        \begin{minipage}{0.22\linewidth}
            \centering\includegraphics[width=\linewidth]{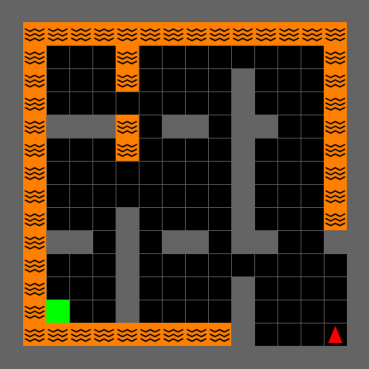}
        \end{minipage}
        \label{fig.minigrid_env_h_b}
    }
    \subfigure[U-Maze]{
        \begin{minipage}{0.22\linewidth}
            \centering\includegraphics[width=\linewidth]{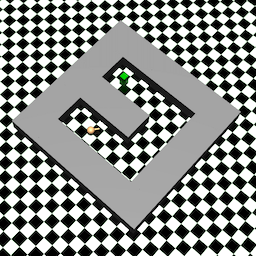}
        \end{minipage}
        \label{fig.u_maze}
    }
    \caption{Environment visualizations. For \texttt{MiniGrid} (\ref{fig.minigrid_env_e_a}-\ref{fig.minigrid_env_h_b}), the context is the position of the goal and key. For \texttt{U-Maze} (\ref{fig.u_maze}) tasks, the context is the position of the goal.}
    \label{fig.environments}
\end{figure}

\textbf{Environments.} The experiments are conducted in two configurable context spaces, as shown in Figure \ref{fig.environments}. The \texttt{MiniGrid} environment is used to evaluate task representation learning and the performance of our method. It is a widely used domain in RL research due to its configurable layouts, partially observable image inputs, exploration challenges, and sparse rewards. The context space includes various tasks that share the same action space but differ in environment layout, object positions, constraints, and reward functions. We designed eight environments with varying levels of difficulty. In the \texttt{U-Maze} environment, both the action space and the context space are continuous, allowing us to study our method in a continuous context setting.

\textbf{Evaluation Metric.} To evaluate the effectiveness of our method, we measure an agent’s learning progress on the target task. Specifically, we compare the asymptotic performance of the algorithm. Additionally, the learning curves include training steps accumulated across all source tasks, reflecting learning efficiency. For the reinforcement learning (RL) algorithm, we use PPO \cite{schulman_proximal_2017} from the \texttt{Stable-Baselines3} library \cite{stable-baselines3} as the base implementation.

The baselines for our evaluation include: 
\begin{itemize}
    \item Default (without curriculum): training directly on the target task using PPO.
    \item Random: sampling the contexts uniformly in the context space.
    \item ALP-GMM \cite{portelas_teacher_2020}: ALP-GMM fits a GMM to past task parameters and their ALP (absolute learning progress) values, sampling new task parameters from high-ALP regions. ALP is computed as the reward difference of the nearest neighbor, and the GMM updates periodically with adaptive Gaussian components while maintaining some random exploration.
    \item Goal GAN \cite{florensa_automatic_2018}: generating contexts of intermediate levels of difficulty for the agent using GAN \cite{goodfellow_generative_2014} and automatically exploring the goal space based on the agent's learning progress.
    \item VDS \cite{zhang_automatic_2020}: generating a goal-sampling curriculum by using the epistemic uncertainty of an ensemble of Q-functions, where higher uncertainty indicates goals at the policy's knowledge frontier.
    \item PLR \cite{jiang_prioritized_2021}: PLR samples the next training level—an instance of a procedural content generation environment—by prioritizing those with higher estimated learning potential when revisited in the future. It maintains a dynamic replay distribution based on level scores and sampling recency, balancing revisiting past levels with exploring new ones to enhance learning efficiency.
    \item CURROT \cite{klink_curriculum_2022}: generating a curriculum in CRL by replacing KL divergence with Wasserstein distance to better account for task similarity and enforcing a minimum performance threshold. It maintains two context buffers for successful and unsuccessful tasks, updating them to ensure the curriculum focuses on tasks at the boundary of agent capability, balancing exploration and targeted learning.
\end{itemize}

\subsection{Tasks with Discrete Action Spaces}
\begin{figure}[t]
    \centering
    \includegraphics[width=\linewidth]{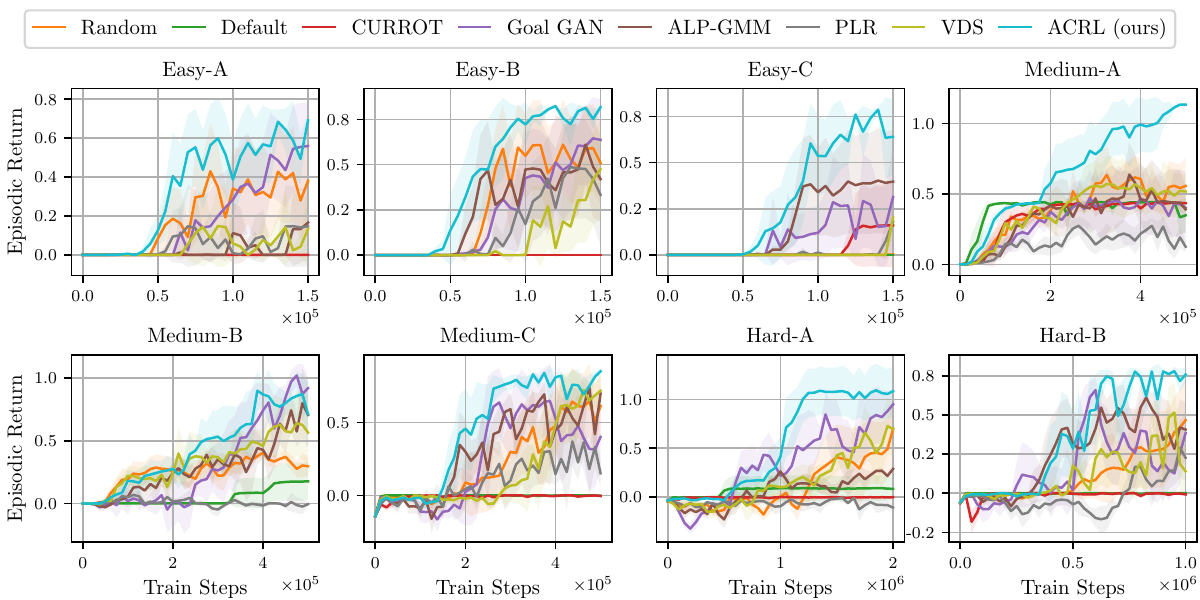}
    \caption{The learning curves in the \texttt{MiniGrid} environments.}
    \label{fig.minigrid_performance}
\end{figure}

\subsubsection{Environment settings}
In the \texttt{MiniGrid} environment, the observation is a \textbf{147-dimensional vector encoding partially observable image-like information}. The agent navigates a maze containing various objects and must complete diverse, complex objectives, such as picking up a key, opening a door, and avoiding obstacles like lava. Because walls block direct paths, the Euclidean distance in the context space no longer reflects task similarity. To ensure consistency, we fix the agent’s initial position and orientation.

The limited observation requires the agent to infer global state information from local perception, increasing the difficulty of long-term planning. Moreover, the long-horizon setting combined with sparse rewards makes learning an effective policy particularly challenging. We categorize the \texttt{MiniGrid} domain into three difficulty levels: \textbf{easy}, \textbf{medium}, and \textbf{hard}.  

\begin{itemize}
    \item In the \textbf{easy} setting, the maximum episode length is small, and the maze layout is simple with no lava.
    \item In the \textbf{medium} setting, the maximum episode length is the same as in easy, but the maze is larger and includes more complex tasks, such as picking up keys and opening doors. Additionally, the environment may contain traps unrelated to the goal to increase exploration difficulty.
    \item In the \textbf{hard} setting, the maze is larger and contains more lava and traps, further hindering exploration.
\end{itemize}

Notably, because tasks involve sequential actions (e.g., picking up keys and unlocking doors), the agent must explore a diverse set of behaviors, making this setting \textbf{substantially more difficult and intricate than simple navigation}. Furthermore, stepping on lava immediately terminates the episode, further limiting the agent’s exploration.

\textbf{State, Action and Context Spaces.} The observation space consists of a $7 \times 7$ grid of tiles in front of the agent. Each tile is represented by a 3-tuple indicating object type, color, and state, resulting in an observation dimensionality of 147 ($7 \times 7 \times 3 = 147$). For door objects, the state takes values 0 (open), 1 (closed), or 2 (locked); all other object types have state 0. Walls (gray areas) block visibility, and the observation is limited to the unobstructed area within the walls. The action space includes seven possible actions, but we use only six. The agent can turn left or right. If the cell directly in front is empty and the agent selects the \textit{forward} action, it moves into that cell; otherwise, it remains in place. To pick up a key, the agent must execute the \textit{pickup} action. To open a door, the agent must use the \textit{toggle} action on a door whose color matches the key it holds. In target-only environments, the context is 2-dimensional, representing the target position. In environments with keys, the context is 4-dimensional, encoding the positions of both the target and the key. Detailed environment specifications are provided in Section \ref{sec.A}.

\textbf{Rewards.} If the goal is not achieved, the agent receives a reward of zero for any action, with a few exceptions. When the goal is reached, the reward depends on the episode length. Specifically, if the episode length is $n$ and the maximum episode length is $H$, the reward for reaching the goal is $1 - 0.5 \frac{n}{H}$. If the agent picks up a key at step $s$, it receives a reward of $0.5(1 - 0.5\frac{s}{H})$. When the agent opens a door, it receives an additional bonus of 0.25. As in \cite{fang_adaptive_2020}, if the agent hits a closed door or a lava region, the episode terminates immediately with a penalty of -0.5.

\textbf{Experiment Details.} Our method can choose any form of distribution, including non-parametric forms, but here we use a Gaussian distribution as the source distribution. For example, the source distributions of \texttt{Easy-A} and \texttt{Medium-A} are set to Gaussians centered at $[2, 3]$ and $[4, 5, 3, 3]$ with variance $[0.5^2, 0.5^2]$ and $[0.5^2, 0.5^2, 0.5^2, 0.5^2]$. Since CURROT requires an explicitly specified initial distribution, we set it to match our source distribution. We set the sampling ratio $\lambda$ to 0.25. Additionally, some tasks may be infeasible. For example, when a locked door must be passed to obtain a key, or when an object is placed inside a wall or at the agent’s initial position. For such cases, we apply rejection sampling to all baselines to ensure only feasible tasks are used. Detailed settings can be found in \ref{sec.A}.

\subsubsection{Results and Analysis}
\begin{figure}[htb]
    \centering
    \subfigure[ACRL (ours)]{
        \begin{minipage}[t]{\linewidth}
            \centering\includegraphics[width=\linewidth]{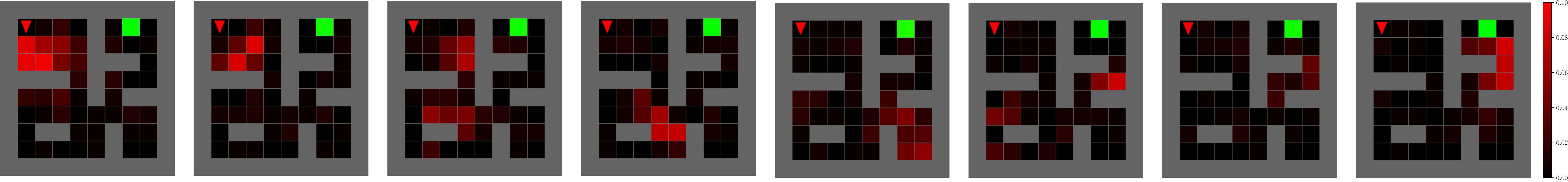}
        \end{minipage}
    }
    \subfigure[CURROT]{
        \begin{minipage}[t]{\linewidth}
            \centering\includegraphics[width=\linewidth]{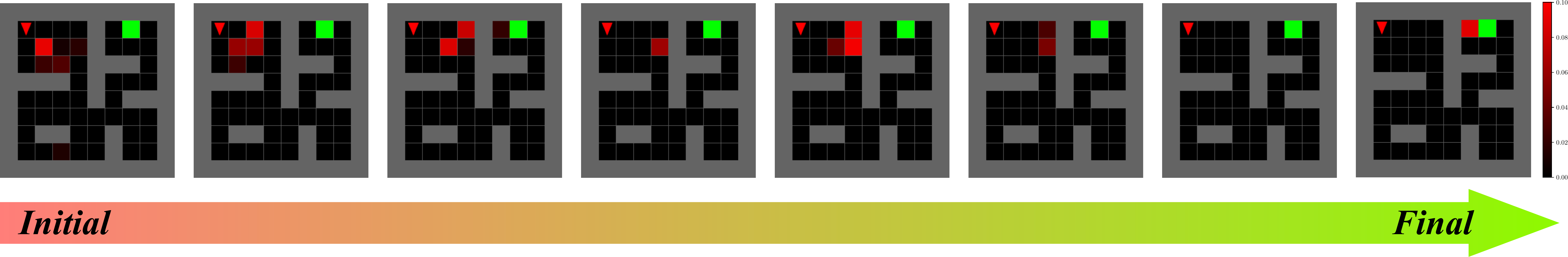}
        \end{minipage}
    }
    \caption{Visualizations of curriculum on \texttt{MiniGrid-Easy-B}. Red rectangles represent a high probability of the location as a target, black indicates a low probability, and the green rectangle represent the target. From left to right, the ACRL-generated distribution gradually moves from the initial position to the gap and gradually moves closer to the target. In contrast, CURROT generates an intermediate distribution that gradually moves closer to the wall to the point where it is difficult to learn the target task.}
    \label{fig.minigrid_curriculum}
\end{figure}

In each task, the experiment results are averaged over $5$ runs with different seeds. The results are shown in Figure \ref{fig.minigrid_performance}. ACRL outperforms the baseline methods in terms of time-to-threshold performance (return $\ge 0.75$) on the target task. Moreover, our method is more stable than other baselines after reaching the threshold due to the target guidance. Since methods such as PLR, Goal GAN, and ALP-GMM are not designed with the target task in mind, the learning efficiency on the target task is unstable. As for interpolation-based methods such as CURROT, since the Euclidean distance metric is not satisfied in the context space, the intermediate task distributions generated may not be adequate for the agent's learning (it may directly hit the wall in the middle of the process and thus be difficult to transfer). In the relatively simple \texttt{Easy} environments, Random instead achieves better performance than harder settings due to the smaller number of samples in the context space.

As shown in Figure \ref{fig.minigrid_curriculum}, in the initial stage, tasks are concentrated near the initial position, which facilitates the agent to quickly learn the policy to goals at critical positions. There are also a few targets distributed in other distant positions, which guide the agent to explore the surrounding environment, enabling the VAE to learn task representations beyond the initial distribution and support prediction. As the VAE gradually learns the representation of easy tasks, the task decoder is synchronously trained and the loss of reduction to context space decreases. The generated task distribution then gradually shifts toward the lower gap, indicating that tasks in this region are considered more difficult. Finally, the distribution progressively approaches the target. This shows that the generated task distribution is shaped by the geometric structure of the environment, guiding the agent along a desired path and facilitating smoother transfer learning. In contrast, CURROT generates interpolation distributions based on rewards associated with contexts, often directing the agent toward goals blocked by obstacles, which hinders effective learning guidance.

\begin{figure}[]
    \centering
    \subfigure[]{
        \begin{minipage}[t]{\linewidth}
            \centering\includegraphics[width=\linewidth]{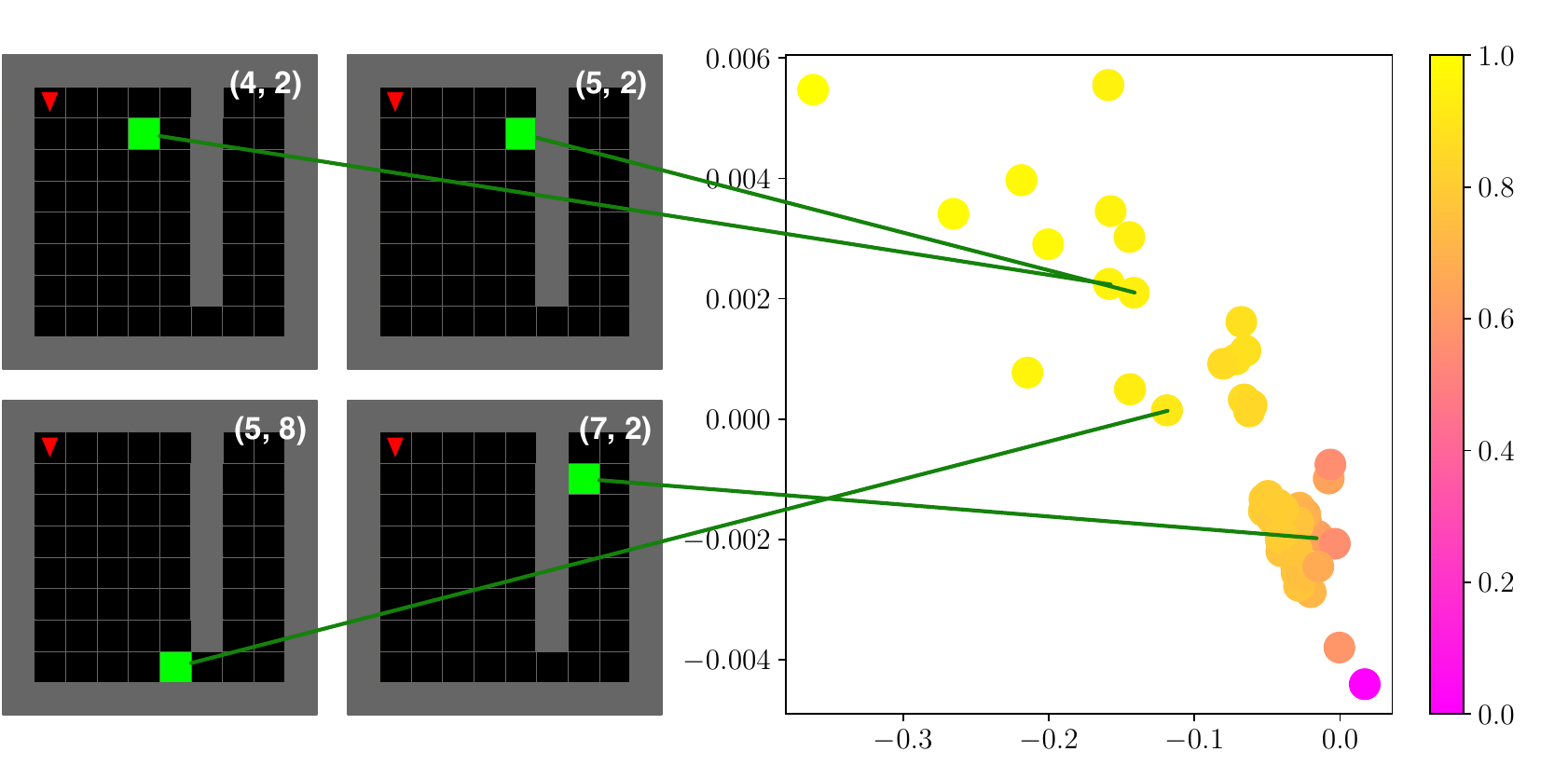}
            \label{fig.minigrid_representation.a}
        \end{minipage}
    }
    \subfigure[]{
        \begin{minipage}[t]{\linewidth}
            \centering\includegraphics[width=\linewidth]{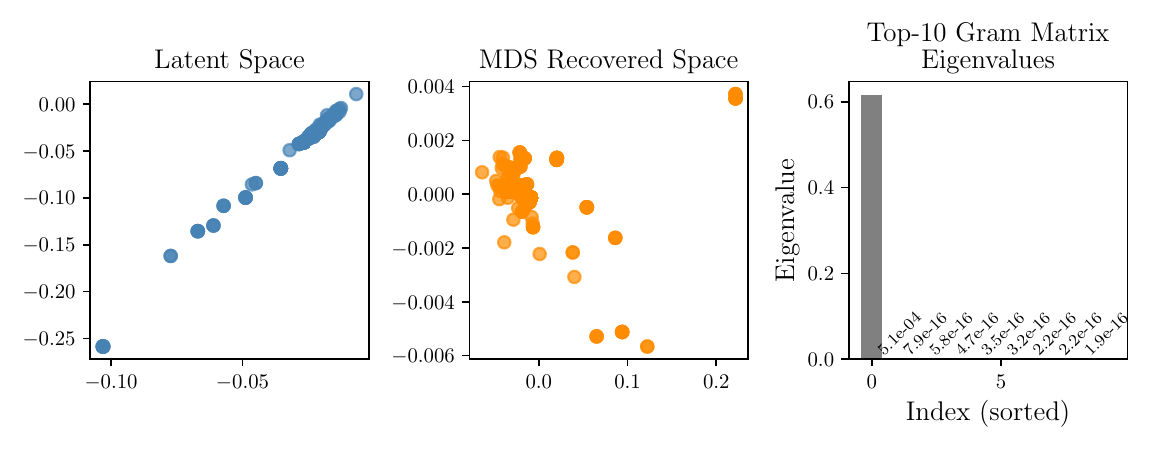}
            \label{fig.minigrid_representation.b}
        \end{minipage}
    }
    \caption{Representation learning results in \texttt{MiniGrid-Easy-A}. (a) Dot positions in the right panel indicate the mean values of the latent space variables, and the colors indicate the episodic return for the corresponding task under the policy. The left panel shows the four tasks configured by $[4, 2]$, $[5, 2]$, $[5, 8]$ and $[7, 2]$, respectively. (b) MDS analysis result. Only top-2 eigenvalues of Gram matrix are positive and others are nearly zero, indicating that a Euclidean metric has been learned.}
    \label{fig.minigrid_representation}
\end{figure}

\subsubsection{Representation Results}
 The correspondence between the output of the latent space and the tasks reflects the fact that the latent space variables can reflect the similarity of the tasks, i.e., information about the metrics of non-Euclidean distances is learned. Figure \ref{fig.minigrid_representation.a} shows the result of task representation. $[5, 2]$ and $[7, 2]$ are closer in the context space, while task $[5, 2]$ and task $[5, 8]$ are farther apart. However, it is clear that task $[7, 2]$ is more difficult than $[5, 8]$ because of the need to go through the gap between the wall and the impossibility to observe on the left side of the wall. Therefore, $[5, 6]$ should be closer to $[5, 2]$ than $[7, 2]$ in terms of similarity, which is illustrated in the results.

 We further conducted a Multidimensional Scaling (MDS) analysis on the latent space, as shown in Figure \ref{fig.minigrid_representation.b}. We obtained the Gram matrix of task embeddings in the VAE buffer and performed spectral decomposition. The results show that only two eigenvalues are significantly positive, while all remaining eigenvalues are close to zero. This indicates that the data effectively lie in a two-dimensional Euclidean space, with the two dominant eigenvalues capturing the intrinsic dimensions of variation. In other words, the latent representation exhibits a manifold structure that can be described in two-dimensional Euclidean geometry. Although our latent representations exhibit approximately Euclidean behavior in practice, this property is emergent rather than assumed, and formal guarantees of Euclidean structure are beyond the scope of this work.

\subsection{Tasks with Continuous Action Spaces}
\begin{figure}[htb]
    \centering
    \includegraphics[width=0.48\linewidth]{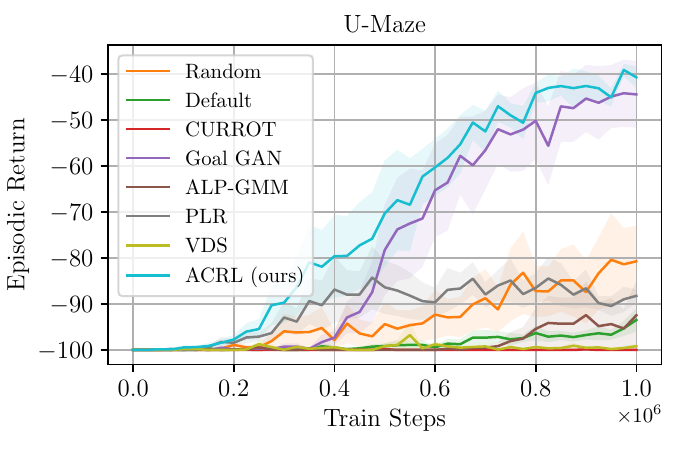}
    \caption{The learning curves in \texttt{U-Maze}. All the curves are averaged over $5$ runs, and the shaded error bars represent the standard variances.}
    \label{fig.u_maze_performance}
\end{figure}

To assess performance in the continuous control task, the \texttt{U-Maze} environment with continuous context space was introduced. The environment is implemented in Mujoco \cite{todorov2012mujoco}. In this environment, the agent must avoid the center barrier, rendering $l_2$ distance as a metric in context space inappropriate. The observations of the agent are the velocity, the position and $5\times 5$ environment information. The context is limited to the range of $[-2, 10]^2$ and we use a 2-dimensional latent space. The source task distribution is a Gaussian distribution centered at $[2, 0]$ with a variance $[0.5^2, 0.5^2]$. The movement range of the agent is limited to the area inside the wall. If the agent does not reach the goal, a penalty of $-1$ is assigned per time step. If the goal is reached (distance $< 1$) then a $+1$ bonus is given and the episode ends. To make exploration more difficult, the maximum number of time steps to reach the goal is 100.

Figure \ref{fig.u_maze_performance} shows that our method achieves better sample efﬁciency compared to baselines as the transfer from the mastered tasks to the new tasks can be relatively smooth. Goal GAN also achieved good asymptotic performance, but time-to-threshold is worse due to the absence of target task. Other baselines perform poorly because the generated tasks may be concentrated in unreachable regions thus reducing the learning efficiency.

\begin{figure}[htb]
    \centering
    \includegraphics[width=\linewidth]{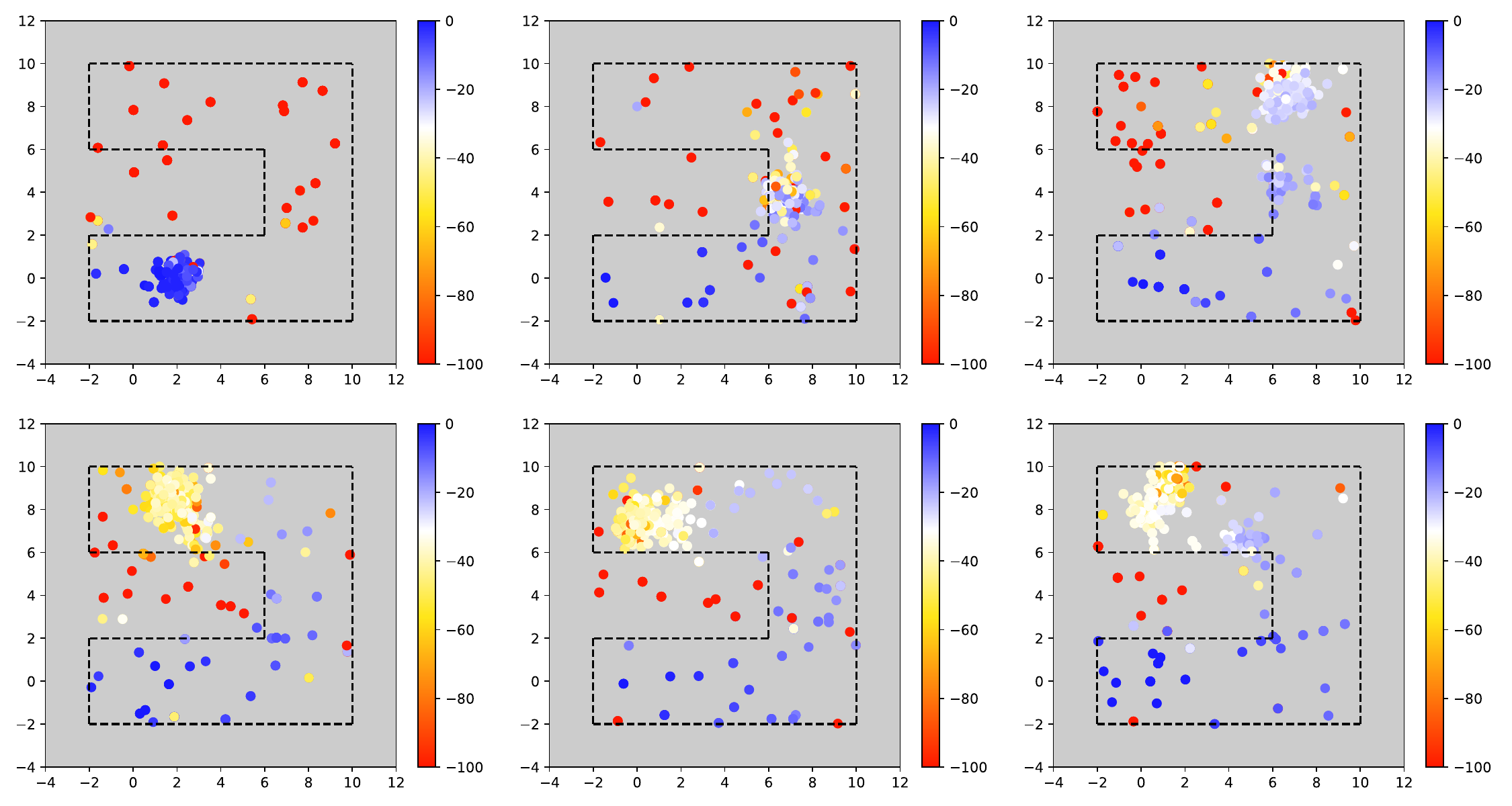}
    \caption{Visualizations of curriculum on \texttt{U-Maze}. The centers of the original source and target are located at $[2, 0]$ and $[0, 8]$, respectively. The distribution gradually transitions from being near the initial position to the right side, ultimately reaching the target goal. In the visual representation, the red color indicates a lower corresponding episodic return, while blue color signifies a higher return.}
    \label{fig.u_maze_curriculum}
\end{figure}

The intermediate task distribution results of our method are shown in Figure \ref{fig.u_maze_curriculum}. In the initial phase, goals are concentrated near the starting point, while goals elsewhere are not able to get a good performance because the policy is not yet fully trained. As training proceeds, the center of the task distribution gradually crosses the middle barrier and moves towards the right side. Eventually, the generated goals aggregate to the target's location, and the return of the target reaches the expectation. Due to a portion of uniform sampling, a small number of exploratory samples were also available at the barriers, but the episodic return is always low.

\subsection{Ablation Study}
\begin{figure}[htb]
    \centering
    \subfigure[\texttt{MiniGrid-Medium-A}]{
        \begin{minipage}{0.47\textwidth}
            \centering\includegraphics[width=\linewidth]{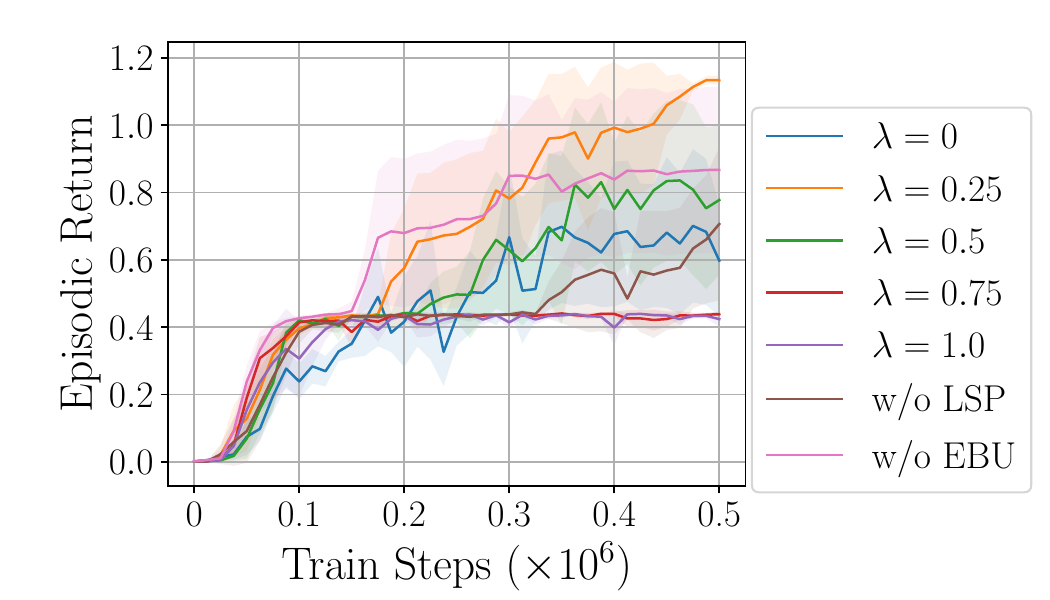}
        \end{minipage}
        \label{fig.ablation_minigrid}
    }
    \subfigure[\texttt{U-Maze}]{
        \begin{minipage}{0.47\textwidth}
            \centering\includegraphics[width=\linewidth]{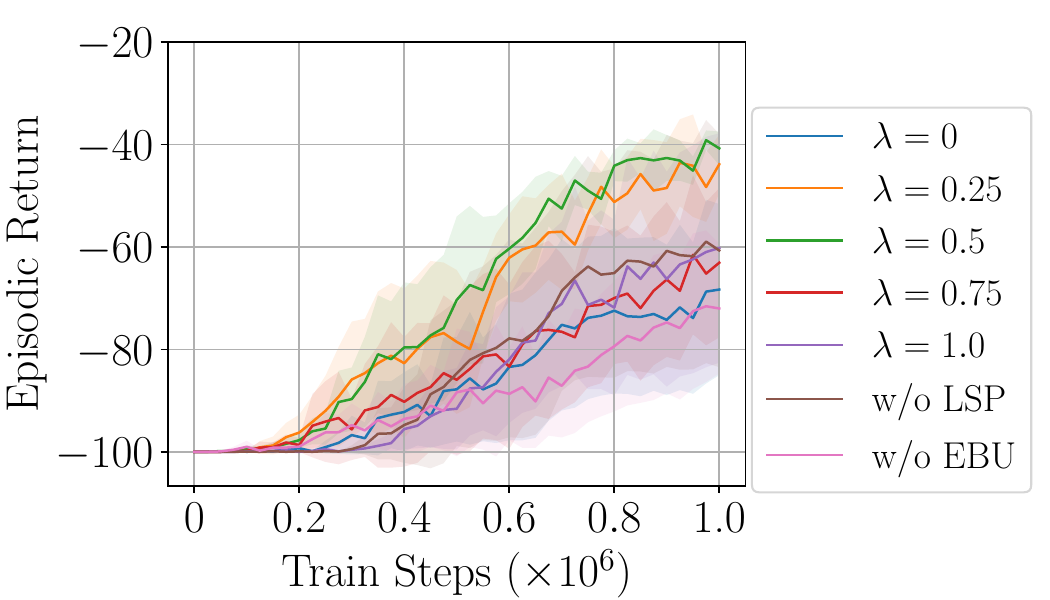}
        \end{minipage}
        \label{fig.ablation_u_maze}
    }
    \caption{Ablation studies on the sampling ratio $\lambda$.}
    \label{fig.ablation}
\end{figure}

\subsubsection{Sampling Ratio $\lambda$}
In ACRL, the updated contexts are stored in the buffer. The ratio $\lambda$ determines the proportion of context predicted in the latent space. When $\lambda$ is too large, the proportion accounted for by uniform sampling is low and thus reduce exploration. Due to the narrow distribution of contexts for representation learning, the loss of the task decoder is relatively large and the prediction accuracy is poor. At the same time, forgetting goals that have already been learned is more likely to happen. In this case the updated contexts are almost indistinguishable from the pre-update ones and the algorithm may fail. However, if $\lambda$ is too small, the algorithm degrades to almost uniform sampling and loses the ability to update the task distribution, and it also reduces the effectiveness of learning.

We conduct ablation studies on \texttt{MiniGrid-Medium-A} and \texttt{U-Maze} environments and analyze the effect of different sampling ratio $\lambda$, which is shown in figure \ref{fig.ablation}. Each curve is averaged over 5 seeds. As can be seen, learning slows down when the update probability is too high, the algorithm performance at $\lambda$ of 1.0 or 0.75 is much worse than that of 0.5 and 0.25. It is nearly equivalent to comparing with uniform sampling when the $\lambda$ is 0. When the update probability is set to 0.25, the sampling efficiency is higher and the algorithm is relatively more stable after convergence with relatively higher average return. Overall, the performance of ACRL receives a large impact when the update probability is too large, the learning process becomes slower and the performance will be relatively poor. Finally, if only LSP or EBU updates are used, the learned representation may not be stable, leading to greater eventual fluctuations while reducing efficiency. We can see that EBU has a greater impact because of the importance of exploration.

\begin{figure}[]
    \centering
    \subfigure[Impact of return threshold $\delta$]{
        \begin{minipage}[t]{\linewidth}
            \centering\includegraphics[width=0.47\linewidth]{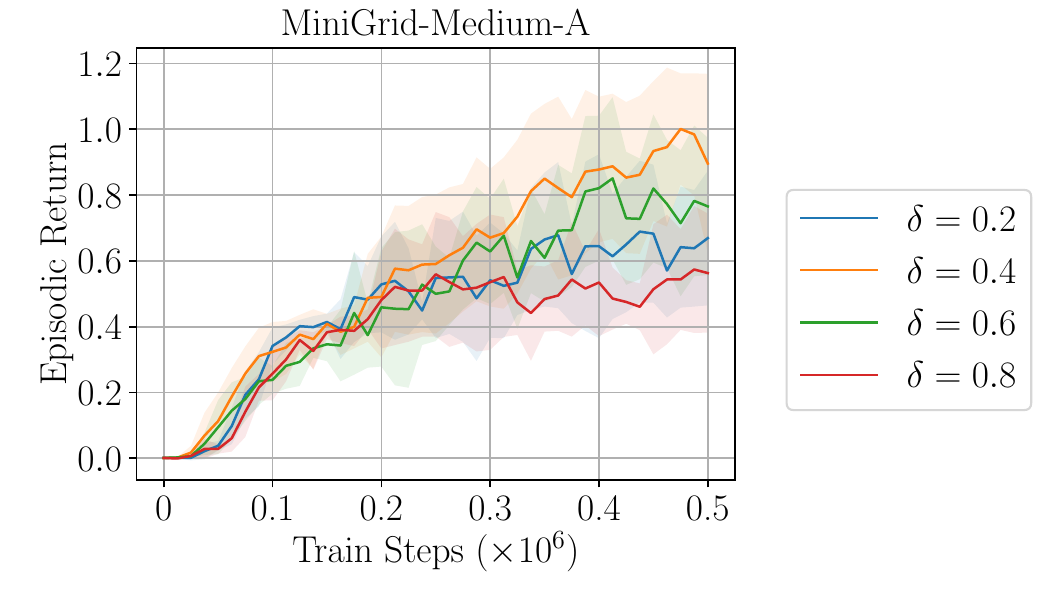}
            \centering\includegraphics[width=0.47\linewidth]{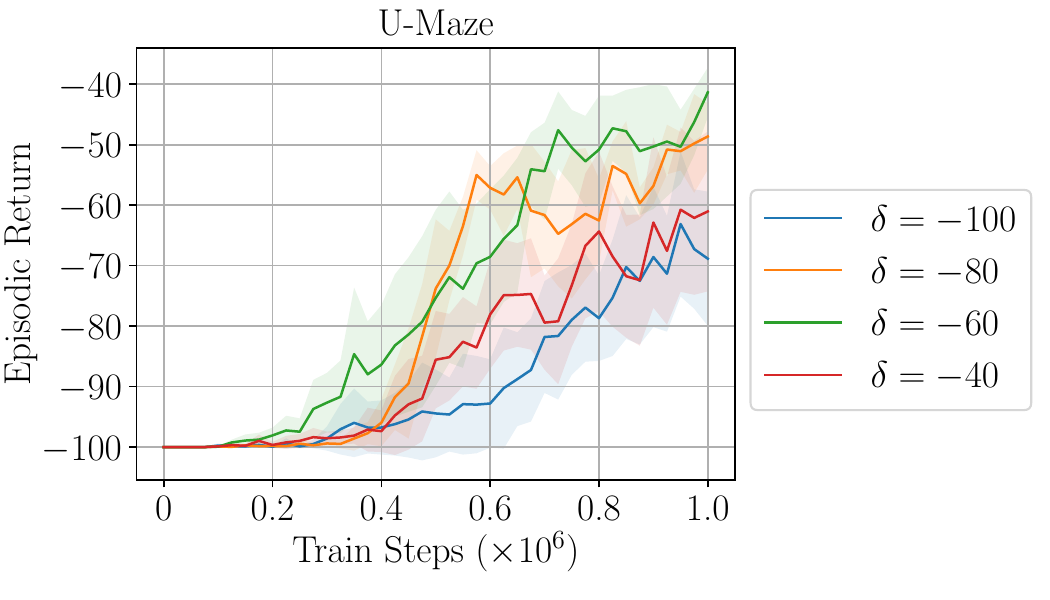}
        \end{minipage}
    }
    \subfigure[Impact of target threshold $\delta_{\text{tar}}$]{
        \begin{minipage}[t]{\linewidth}
            \centering\includegraphics[width=0.47\linewidth]{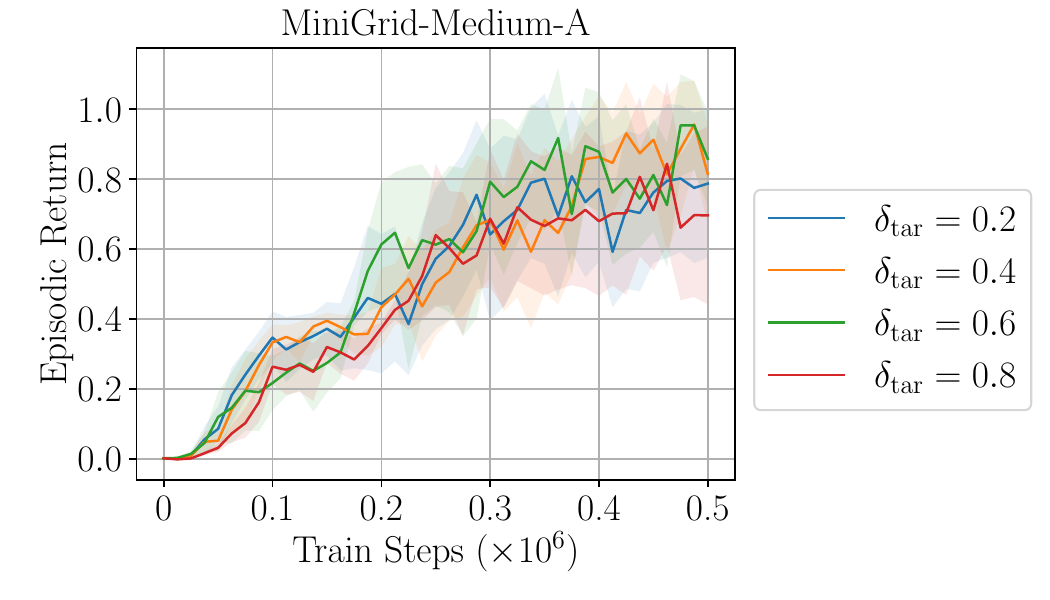}
            \centering\includegraphics[width=0.47\linewidth]{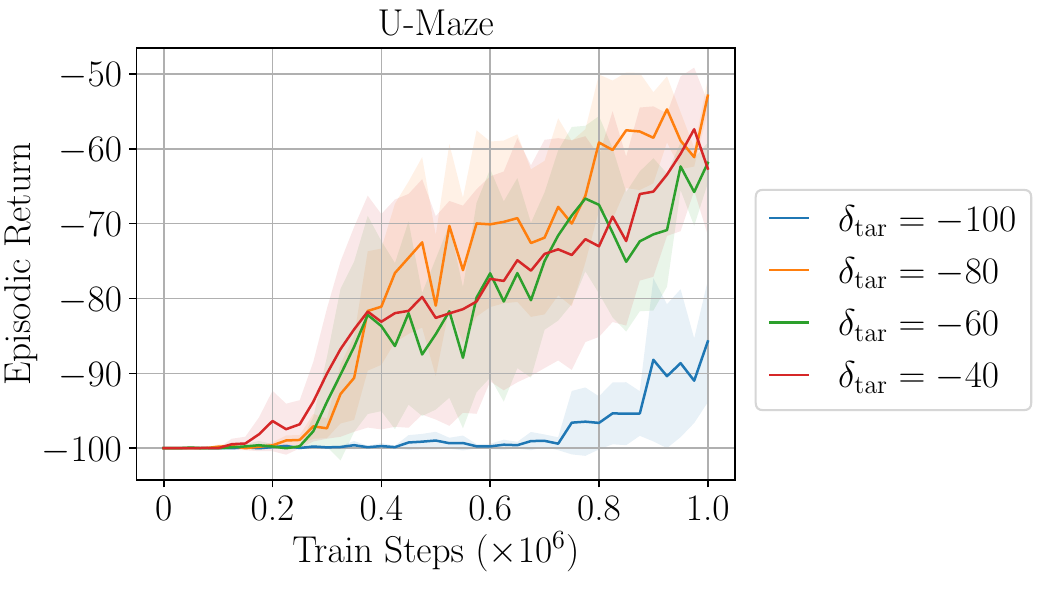}
        \end{minipage}
    }
    \subfigure[Impact of update threshold $G$]{
        \begin{minipage}[t]{\linewidth}
            \centering\includegraphics[width=0.47\linewidth]{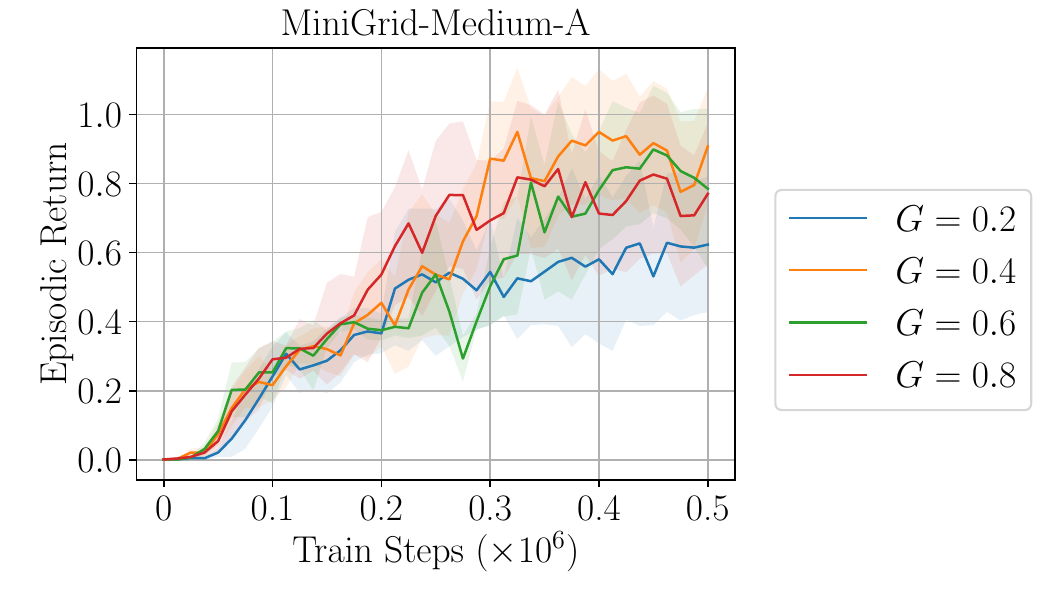}
            \centering\includegraphics[width=0.47\linewidth]{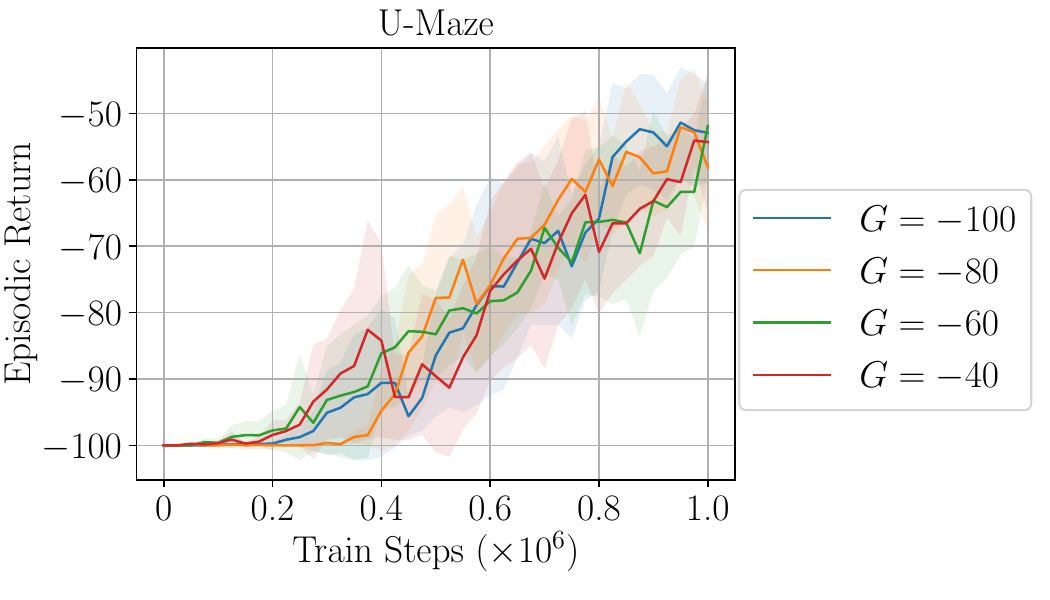}
        \end{minipage}
    }
    \caption{Impact of curriculum parameters.}
    \label{fig.ablation_curriculum_parameters}
\end{figure}

\subsubsection{Curriculum Parameters}
To assess the robustness of our approach, we further study the sensitivity of key curriculum parameters, including the return threshold $\delta$, the update threshold $\delta_{\text{tar]}}$, and the target return threshold $G$. We vary each parameter within a reasonable range and evaluate performance on both \texttt{MiniGrid} and \texttt{U-Maze}. As shown in Figure \ref{fig.ablation_curriculum_parameters}, the method consistently achieves strong performance, provided that parameters are not set to extreme values. This indicates that the algorithm is stable across a wide range of parameter choices.

\subsubsection{Decoders of Task Representation Learning}
\begin{figure}[]
    \centering
    \subfigure[\texttt{MiniGrid-Medium-C}]{
        \begin{minipage}[t]{0.47\linewidth}
            \centering\includegraphics[width=\linewidth]{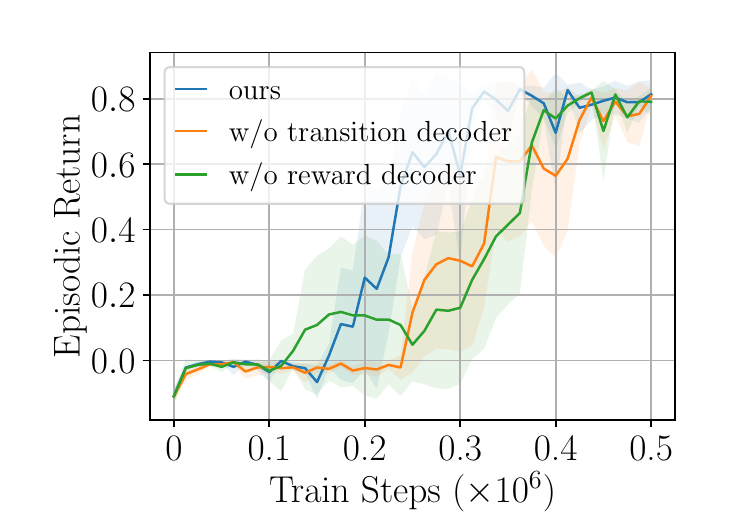}
        \end{minipage}
    }
    \subfigure[\texttt{U-Maze}]{
        \begin{minipage}[t]{0.47\linewidth}
            \centering\includegraphics[width=\linewidth]{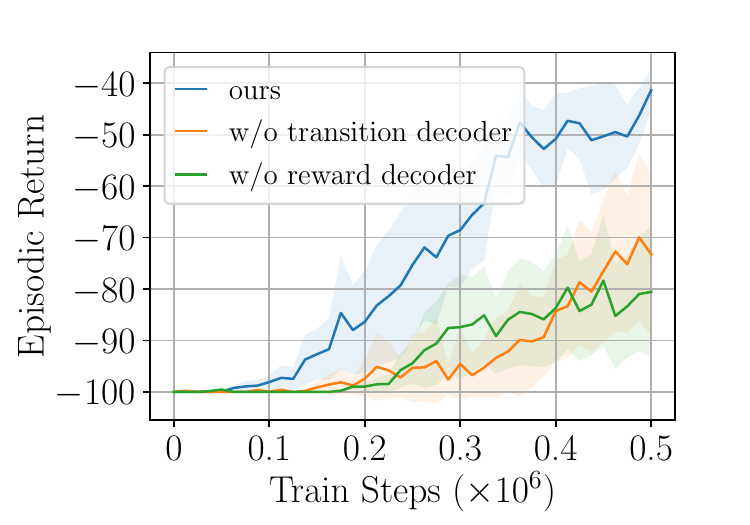}
        \end{minipage}
    }
    \caption{Ablation studies on decoders of task representaion learning.}
    \label{fig.ablation_decoders}
\end{figure}

We further investigate the contribution of different decoders through ablation. The reward and transition decoders are integral to the VAE objective, ensuring that the latent embeddings capture both reward-relevant and transition-relevant features. Without these components, the learned representations degenerate, leading to poor task discrimination and unstable updates. To validate this, we conducted ablation experiments where either the reward decoder or the dynamics decoder was removed. As shown in Figure \ref{fig.ablation_decoders}, in both cases, performance degraded, especially in \texttt{U-Maze} domain. Furthermore, eliminating either component impaired the quality of the latent space and reduced the overall learning performance. This indicates that both decoders play comparable roles in representation learning.

\section{Conclusion}\label{sec.conclusion}
In this work, we introduce ACRL, a novel curriculum RL framework designed to generate the intermediate tasks according to the policy learning process and accelerate the learning efficiency on the target task. Leveraging the capabilities of the task representation learning, we encode the trajectories sampled by the policy for specific tasks to obtain task representations with measurable task similarity. With the task representations, we can estimate the similarity between tasks and exploration bound of the policy naturally, which can derive smooth intermediate task distributions and enable the approach to handle non-Euclidean metric task spaces. Our method notably outperforms baselines that do not consider the target task, showcasing superior performance. Empirical evidence substantiates the advantages of our proposed method, emphasizing the consideration of the measurable task similarity.

Although it has demonstrated promising results, a current limitation of ACRL is its reliance on parametric context representations. The method cannot directly handle non-parametric task definitions such as symbolic or language-based contexts, though future work may extend ACRL by incorporating semantic encoders.

\section*{Acknowledgments}
We acknowledge funding in support of this work from the Project supported by the Key Program of the National Natural Science Foundation of China (Grant No.62306088), the Natural Science Foundation of Heilongjiang Province (Grant No.YQ2024F007), 2024QNRC001 (NO.YESS 20240415) and Songjiang Lab (Grant No.SL20230309).

\appendix
\section{Experiment Details}\label{sec.A}
We show the hyperparameters of the experiments in Appendix \ref{ssec.hp}. For parameters not mentioned in the tables, use the default parameters or the same parameters as the other experiment setups.

\subsection{Hyperparameters}\label{ssec.hp}
The hyperparameter settings for ACRL are presented in Table \ref{tbl.acrl_hyperparameters}. Additionally, he hyperparameters of baselines used in \texttt{MiniGrid} and \texttt{U-Maze} can be found in Table \ref{tbl.minigrid_hyperparameters} and Table \ref{tbl.u_maze_hyperparameters}, respectively.

\begin{table}
    \caption{\texttt{MiniGrid} ACRL Hyperparameters}
    \centering
    \begin{tabular}{ll}
    \toprule
    hyperparameter & value \\
    \midrule
    VAE learning rate & 0.005 \\
    state\_embedding\_size & 64 \\
    action\_embedding\_size & 8 \\
    reward\_embedding\_size & 8 \\
    vae\_buffer\_size & 256 \\
    batch\_size & 32 \\
    task buffer size & 256 \\
    target samples $k$ & 10 \\
    return threshold $\delta$ & 0.4 \\
    update threshold $G$ & 0.5 \\
    target threshold $\delta_{\text{tar}}$ & 0.4 \\
    EBU noise $\sigma$ & 1.0 \\
    exponential distribution $\beta$ & 1.0 \\
    step size $\alpha$ & 0.9 \\
    encoder network architecture & MLP(128, 128) \\
    reward decoder network architecture & MLP(64, 64) \\
    state transition decoder network architecture & MLP(128, 128) \\
    task decoder network architecture & MLP(128, 128) \\
    \bottomrule
    \end{tabular}
    \label{tbl.acrl_hyperparameters}
\end{table}

\begin{table}
    \caption{\texttt{MiniGrid} Baselines Hyperparameters}
    \centering
    \begin{tabular}{ll}
    \toprule
    PPO & value \\
    \midrule
    gamma & 0.95 \\
    net architecture & $[128, 128, 128]$ \\
    activation\_fn & Tanh \\
    batch\_size & 128 \\
    gae\_lambda & 0.99 \\
    \midrule
    VDS & value \\
    \midrule
    num\_Q & 5 \\
    learning rate & $10^{-3}$ \\
    num\_epoch & 3 \\
    \midrule
    ALP-GMM & value \\
    \midrule
    $p_{rnd}$ & 0.1 \\
    fitting rate $N$ & 100 \\
    max nunber of Gaussians $k_{max}$ & 500 \\
    \midrule
    Goal GAN & value \\
    \midrule
    noise level & 0.05 \\
    fit rate & 200 \\
    $p_{old}$ & 0.05 \\
    \midrule
    PLR & value \\
    \midrule
    replay rate & 0.85 \\
    buffer size & 100 \\
    $\beta$ & 0.45 \\
    $\rho$ & 0.15 \\
    \midrule
    CURROT & value \\
    \midrule
    $\delta$ & 0.3 \\
    metrics $\epsilon$ & 1.0 \\
    number of episodes between updates & 40 \\
    \bottomrule
    \end{tabular}
    \label{tbl.minigrid_hyperparameters}
\end{table}

\begin{table}
    \caption{\texttt{U-Maze} Hyperparameters}
    \centering
    \begin{tabular}{ll}
    \toprule
    PPO & value \\
    \midrule
    gamma & 0.99 \\
    \midrule
    ALP-GMM & value \\
    \midrule
    $p_{rnd}$ & 0.1 \\
    fitting rate $N$ & 100 \\
    max nunber of Gaussians $k_{max}$ & 1000 \\
    \midrule
    Goal GAN & value \\
    \midrule
    noise level & 0.1 \\
    fit rate & 200 \\
    $p_{old}$ & 0.2 \\
    \midrule
    PLR & value \\
    \midrule
    replay rate & 0.95 \\
    buffer size & 100 \\
    $\beta$ & 0.1 \\
    $\rho$ & 0.3 \\
    \midrule
    ACRL & value \\
    \midrule
    return threshold $\delta$ & -70 \\
    update threshold $G$ & -60 \\
    target threshold $\delta_{\text{tar}}$ & -40 \\
    task buffer size & 128 \\
    \bottomrule
    \end{tabular}
    \label{tbl.u_maze_hyperparameters}
\end{table}

\subsection{Environment Details}\label{ssec.environmentt_details}
In Table \ref{tbl.minigrid_details}, we provide the maximum number of steps for each task setting, the center of the initial distribution, and the update frequency.

\begin{table}
    \caption{\texttt{MiniGrid} domain details}
    \centering
    \begin{tabular}{lcccc}
    \toprule
    & max steps & initial center & update frequence & latent dim \\
    \midrule
    Easy-A & 50 & [2, 3] & 500 & 2 \\
    Easy-B & 75 & [2, 3] & 500 & 2 \\
    Easy-C & 75 & [2, 3] & 500 & 2 \\
    Med-A & 50 & [4, 5, 3, 3] & 500 & 4 \\                           
    Med-B & 75 & [8, 3, 4, 1] & 500 & 4 \\
	Med-C & 100 & [2, 6] & 500 & 2 \\
	Hard-A & 200 & [4, 5, 3, 3] & 500 & 4 \\
	Hard-B & 75 & [13, 11] & 250 & 2 \\
	\bottomrule
    \end{tabular}
    \label{tbl.minigrid_details}
\end{table}
\clearpage

\bibliographystyle{elsarticle-harv} 
\bibliography{bibliography}






\end{document}